\documentclass[lettersize,journal]{IEEEtran}
\usepackage{amsmath,amsfonts}

\usepackage[ruled,linesnumbered]{algorithm2e}
\usepackage{array}
\usepackage{textcomp}
\usepackage{stfloats}
\usepackage{url}
\usepackage{verbatim}
\usepackage{graphicx}
\usepackage{cite}
\usepackage{verbatim}
\usepackage{graphicx}
\usepackage{textcomp}
\usepackage{setspace}
\usepackage{subfigure}
\usepackage{subfiles}  
\usepackage{mathrsfs}
\usepackage{xcolor}
\usepackage[numbers,sort&compress]{natbib}
\newcommand{\name}{SafeSign~}
\newcommand{\sname}{SafeSign}

\begin{document}

\title{Secure Traffic Sign Recognition: An Attention-Enabled Universal Image Inpainting Mechanism against Light Patch Attacks
}

\author{Hangcheng Cao,~
        Longzhi Yuan,~
        Guowen Xu,~
        Ziyang He,~
        Zhengru Fang,~
        and Yuguang Fang,~\IEEEmembership{Fellow,~IEEE}
}        

\maketitle

\begin{abstract}
Traffic sign recognition systems play a crucial role in assisting drivers to make informed decisions while driving. However, due to the heavy reliance on deep learning technologies, particularly for future connected and autonomous driving, these systems are susceptible to adversarial attacks that pose significant safety risks to both personal and public transportation. Notably, researchers recently identified a new attack vector to deceive sign recognition systems: projecting well-designed adversarial light patches onto traffic signs. In comparison with traditional adversarial stickers or graffiti, these emerging light patches exhibit heightened aggression due to their ease of implementation and outstanding stealthiness. To effectively counter this security threat, we propose a universal image inpainting mechanism, namely, \sname. It relies on attention-enabled multi-view image fusion to repair traffic signs contaminated by adversarial light patches, thereby ensuring the accurate sign recognition. Here, we initially explore the fundamental impact of malicious light patches on the local and global feature spaces of authentic traffic signs. Then, we design a binary mask-based U-Net image generation pipeline outputting diverse contaminated sign patterns, to provide our image inpainting model with needed training data. Following this, we develop an attention mechanism-enabled neural network to jointly utilize the complementary information from multi-view images to repair contaminated signs. Finally, extensive experiments are conducted to evaluate \sname's effectiveness in resisting potential light patch-based attacks, bringing an average accuracy improvement of 54.8\% in three widely-used sign recognition models.

\end{abstract}
\thispagestyle{empty}

\section{Introduction}
\label{sec:intro}
Traffic sign recognition (TSR) systems play a crucial role in identifying signs on roads and assisting drivers or control modules in making driving decisions~\cite{meuter2011decision}. In recent years, the performance of TSRs has significantly improved, thanks to the advancements in learning-based computer vision technologies. These systems even outperform human visual systems in many challenging driving environments and scenarios~\cite{krizhevsky2012imagenet,le2011learning}. Given the exceptional performance, emerging vehicle models (e.g., Tesla's Model 3~\cite{tesla} and BMW's I7~\cite{bmw}) incorporate TSRs as crucial components of driving assistance systems. However, security researchers have raised concerns about TSRs due to their reliance on deep learning-driven models, which are vulnerable to adversarial examples~\cite{sato2024invisible,duan2021adversarial,zhong2022shadows,lovisotto2021slap,10555359}. In particular, attackers can create maliciously crafted examples to deceive deep neural networks, thereby misleading sign recognition models to make false predictions. The consequence of successful attacks via adversarial examples is serious. For example, on high-speed lanes, TSR misjudgments compel drivers to spend more time judging traffic information, thus resulting in rear-end collisions or even fatal accidents~\cite{duan2021adversarial,lovisotto2021slap}. In general, such attacks pose huge risks to both personal and public transportation.

Taking into account the differences in attack approaches, current physical adversarial examples against TSRs can be categorized into two main types: contact stickers/graffiti~\cite{wang2023does,hingun2023reap, JiaLZL0022,zhao2019seeing,eykholt2018robust} and non-contact light patches~\cite{lovisotto2021slap,duan2021adversarial,zhong2022shadows,wang2023rfla,sato2024invisible}. The first type involves attackers physically accessing and contaminating target traffic signs. For instance, attackers attach square stickers to sub-regions of signs to change the original feature patterns~\cite{duan2021adversarial}, or modify the color distribution of the entire sign~\cite{JiaLZL0022}. Because of these physical modifications, such attacks can be easily detectable upon simple physical examination. Additionally, the contaminated areas often exhibit noticeable color differences compared to the authentic signs, which can be detected and removed by road maintenance staff. Recently, researchers introduce a new non-contact attack method, that is, remotely projecting well-designed visible or invisible light patches on signs. This approach is more aggressive since it is easy to deploy and holds better stealthiness. Specifically, attacks can be launched by remotely contaminating target signs, rather than through cumbersome contact modifications. Moreover, some patches appear as natural light shadows~\cite{lovisotto2021slap,duan2021adversarial,zhong2022shadows,wang2023rfla} or employ infrared lights invisible to human eyes~\cite{sato2024invisible}, introducing imperceptible sign pattern variations. Once the attack is successfully launched and makes a preconceived malign outcome, attackers can easily deactivate the light sources to eliminate evidence. This emerging non-contact attack obviously poses greater safety risks compared to the contact type. 

\begin{figure}[t]
\centering
\includegraphics[width=0.48\textwidth]{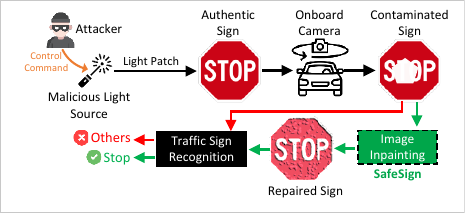}
\caption{Illustration of \sname's role, that is, repairing signs contaminated by adversarial light patches. We take the ``STOP'' sign repair as an example: with enabling \sname, the contaminated sign pattern is repaired and then TSR can correctly recognize its label.}
\label{fig:illSafeSign}
\end{figure}

It is worrisome that current defense mechanisms are unable to effectively mitigate the risk posed by adversarial light patches, as evident in the following aspects: i) They require retraining recognition models according to specific training methods, which is inappropriate in TSR systems due to their closed-loop design manners~\cite{duan2021adversarial,lovisotto2021slap,zhong2022shadows,wang2023rfla}. ii) They solely support protection for particular attack modes, assuming that attack settings and light patch patterns are known in advance~\cite{lovisotto2021slap,duan2021adversarial,zhong2022shadows,wang2023rfla,sato2024invisible}. iii) They necessitate adding new hardware modules on vehicles to filter specific malicious light components, which is challenging to deploy because it requires redesigning existing hardware systems~\cite{sato2024invisible}. The commonality among the above issues highlights that these defense mechanisms are customized solutions for specific attack patches/modes, lacking universality. \textit{Therefore, establishing a universal protection mechanism for TSRs to effectively counter various adversarial light patch-based attacks is still an open problem.} To deal with this issue, we propose an image inpainting mechanism, namely, \sname, which repairs various contaminated sign images before feeding them into recognition models. The function of \sname\ in the overall sign recognition architecture is depicted in Fig.~\ref{fig:illSafeSign}. It is responsible for repairing images contaminated by malicious light patches, thus providing the recognition model with reconstructed signs that can be accurately identified. Our solution is promising, but three critical issues must be properly addressed. Firstly, vehicle-installed TSRs always adhere to a closed-loop design principle. Consequently, during the implementation process, \name cannot make assumptions such as retraining sign recognition models, as done by existing defense strategies. Secondly, existing mechanisms only function when they have prior knowledge of attack modes and patch patterns, allowing them to design customized countermeasures. However, acquiring this prior knowledge is hard, if not impossible, given that attack methods and settings are fully controlled by attackers. Therefore, constructing a defense mechanism capable of effectively countering various attacks is urgently needed. Lastly, as the vehicle moves, onboard cameras continuously capture sign images from multiple views. These images, captured under various lighting conditions, contain distinct information essential for the contaminated sign repair. Therefore, it is necessary to carefully consider how to effectively fuse the information from multi-view images to repair sign images and hence ensure accurate recognition.

We address the above issues by the following three key steps. Firstly, by analyzing the attack launching process, we discover that the most effective countermeasure without retraining existing TSRs is to remove the interference of light patches before feeding sign images into recognition models. To achieve this, we develop an image inpainting mechanism as a preset module for TSRs, enabling the repair of contaminated sign images. Secondly, we investigate the root cause of TSR misjudgments resulting from various light patch-based attacks. Specifically, regardless of attack modes and settings, the essence of these attacks lies in disrupting the global or local feature patterns of authentic signs. Therefore, we design a mask-based U-Net~\cite{ronneberger2015u} adversarial patch generation model that outputs various potential contaminated patterns. This approach provides sufficient samples for our sign reconstruction model to learn how to mitigate the impact of adversarial patches on signs. Lastly, we reveal the contribution difference of sign images captured from different views to the repair process. Based on this understanding, we design an attention mechanism-driven mechanism jointly using SENet network~\cite{hu2018squeeze} and multi-head self-attention module~\cite{xu2021supervised} to fully exploit the complementary information provided by multi-view sign images. We verify the \sname's performance by applying it to common one-stage (i.e., YOLO5~\cite{redmon2016you}) and two-stage (i.e., LeNet~\cite{lecun1998gradient} and GoogleNet~\cite{szegedy2015going}) sign recognition models. The experimental results demonstrate that \name significantly enhances the average recognition accuracy and precision by up to 54.8\% and 58.5\%, respectively, against four representative light patch-based attacks launched by infrared~\cite{sato2024invisible}, laser~\cite{duan2021adversarial}, artificial shadows~\cite{zhong2022shadows}, and projector~\cite{lovisotto2021slap}. In a nutshell, our contributions are summarized as follows.
\begin{itemize}
    \item We propose \sname, a universal mechanism for defending against potential light patch-based attacks by repairing contaminated traffic signs. Its superiority lies in the fact that it does not require retraining existing recognition models, obtaining prior knowledge of attack settings, and modifying the TSR hardware architecture.
   \item We design an U-Net based adversarial sign generation model that combines binary masks to generate contaminated signs, effectively representing potential contaminated patterns of light patches crucial for building the reconstruction model.
    \item We construct an attention mechanism-driven sign reconstruction neural network to effectively utilize information from multi-view images to repair contaminated signs, thereby providing TSRs with more reliable detection.
    \item We conduct extensive experiments on public datasets and widely-used recognition models to evaluate the effectiveness of \sname. The results demonstrate its capability to resist existing light patch-based attacks and effectively ensure the recognition accuracy.
\end{itemize}
Compared with existing defense studies, \name exhibits the following specific advantages: The image inpainting process is independent of attack modes and patch patterns, making \name a universal protection mechanism. It does not require retraining traffic recognition models, thus adhering to the closed-loop design approach. Additionally, since it operates at the algorithm level, \name does not necessitate changes to TSR hardware architectures, ensuring cost-effectiveness.

\section{Related Work}
\label{sec:relatedWork}
We hereby introduce existing adversarial attacks (including light patch-based types) and corresponding defense mechanisms in the context of TSRs. We then highlight the differences between our \name and these existing defense ways.

\subsection{Adversarial Attack Against TSR}
Recent years have witnessed a marked surge in concern over the substantial risks presented by adversarial attacks that targeting deep learning-driven TSR systems~\cite{3485133}. Adversarial attacks follow the basic principle of introducing nearly imperceptible perturbations to the inputs of deep neural networks, thereby altering the final prediction result~\cite{cao2024security}. Research on adversarial examples within TSRs focuses on two aspects: digital systems~\cite{dong2018boosting,goodfellow2014exp,7958570,papernations} and physical systems~\cite{wang2023does,hingun2023reap, JiaLZL0022,zhao2019seeing,eykholt2018robust,lovisotto2021slap,duan2021adversarial,zhong2022shadows,wang2023rfla,sato2024invisible}, respectively. In the digital domain, adversarial perturbations are directly applied to the input data and are typically constrained by $L_p$-norm to ensure that the alterations remain indiscernible. Adversarial attacks can be classified into white-box and black-box categories~\cite{zhong2022shadows}. Compared to white-box attacks, the black-box one does not necessitate detailed knowledge of the internal structure of targeted models. However, when deployed in the real world, digital-level attacks often suffer from poor performance because important parameters such as lighting conditions and camera settings are not taken into account~\cite{duan2021adversarial, sato2024invisible}. Therefore, some studies explore adversarial examples of physical attacks. For example, earlier works involved sticking stickers or graffiti on target signs to contaminate their original patterns~\cite{wang2023does,hingun2023reap, JiaLZL0022,zhao2019seeing,eykholt2018robust}. Although these methods demonstrated the new security risk, the process of launching such attacks was easy to detect or trace. To enhance the attack stealthiness, attackers introduce a new attack vector by strategically applying well-designed light patches to signs~\cite{lovisotto2021slap,duan2021adversarial,zhong2022shadows,wang2023rfla,sato2024invisible}, with the aim of inducing misclassification by the models. For instance, infrared lights~\cite{sato2024invisible} and laser beams~\cite{duan2021adversarial} projected onto traffic signs can lead to incorrect outputs from TSR systems. In general, the emergence of light patches poses a heightened threat to public transportation safety.

\subsection{Defense Mechanism for TSR}
Currently, there is no universal mechanism to address the emerging threats posed by light patch attacks on TSR systems. Existing defense mechanisms are typically designed to address specific attack patterns, which may leave systems vulnerable to alternative attack methods. For example, there have been suggestions of using filters to resist imperceptible infrared patches~\cite{sato2024invisible}, as well as proposals for traffic sign upgrades that install radio frequency units to transmit wireless signs to represent relevant traffic sign information~\cite{3625809}. These approaches require changes to either vehicle or traffic sign hardware, introducing extra costs that hinder widespread implementation. Moreover, some researchers take the approach of understanding known attack patterns and generate corresponding adversarial samples, retraining recognition models to enhance their resistance to attacks~\cite{lovisotto2021slap,duan2021adversarial,zhong2022shadows}. However, this method exposes significant practical limitations. For instance, retraining a model deviates from the closed-loop design principle of TSRs. Assuming that attack patterns are known in advance is unrealistic, as attackers can constantly adapt their methods. To address the above-mentioned issues, we propose a universal image inpainting mechanism to reconstruct signs contaminated by potential light patches. As a preset module, there is no need to retrain recognition models and change existing hardware components. Moreover, its plug-and-play nature allows for wide applicability across TSRs.

\section{Background and Preliminaries}
\label{sec:background}
\subsection{Traffic Sign Recognition}
\label{subsec:tsrType}
\begin{figure}[t]
\centering
\includegraphics[width=0.5\textwidth]{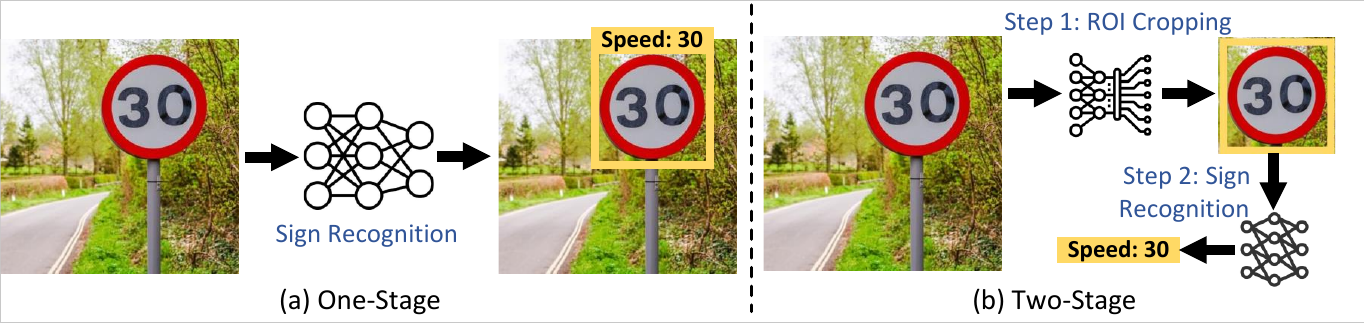}
\caption{Illustration of traffic sign recognition workflows in (a) one-stage and (b) two-stage architectures.}
\label{fig:recWorkflow}
\end{figure}

After obtaining images provided by onboard cameras, TSRs employ deep neural networks to recognize corresponding traffic sign types. These sign recognition networks are primarily divided into single-stage and two-stage architectures depending on their workflow, as illustrated in Fig.~\ref{fig:recWorkflow}. Single-stage models automatically segment an image into multiple sub-regions and simultaneously output the corresponding sign labels. In contrast, two-stage models first use object detectors to identify the position of traffic signs within an image, then crop them into regions of interest (ROI) and perform detailed categorization on these cropped regions. The former has the advantage of concise architecture and low computation overhead, while the latter offers higher accuracy. Despite employing different methodologies and exhibiting respective working characteristics, both architectures aim to recognize target sign regions (i.e., ROIs) in the captured images. Consequently, our study focuses on analyzing potential contamination patterns within these ROIs and repairing them using our sign repairing mechanism.

\subsection{Light Patch-based Attacks}
\begin{figure}[h]
\centering
\subfigure[Infrared]{
\begin{minipage}[t]{0.21\linewidth}
\centering
\includegraphics[width=1\textwidth]{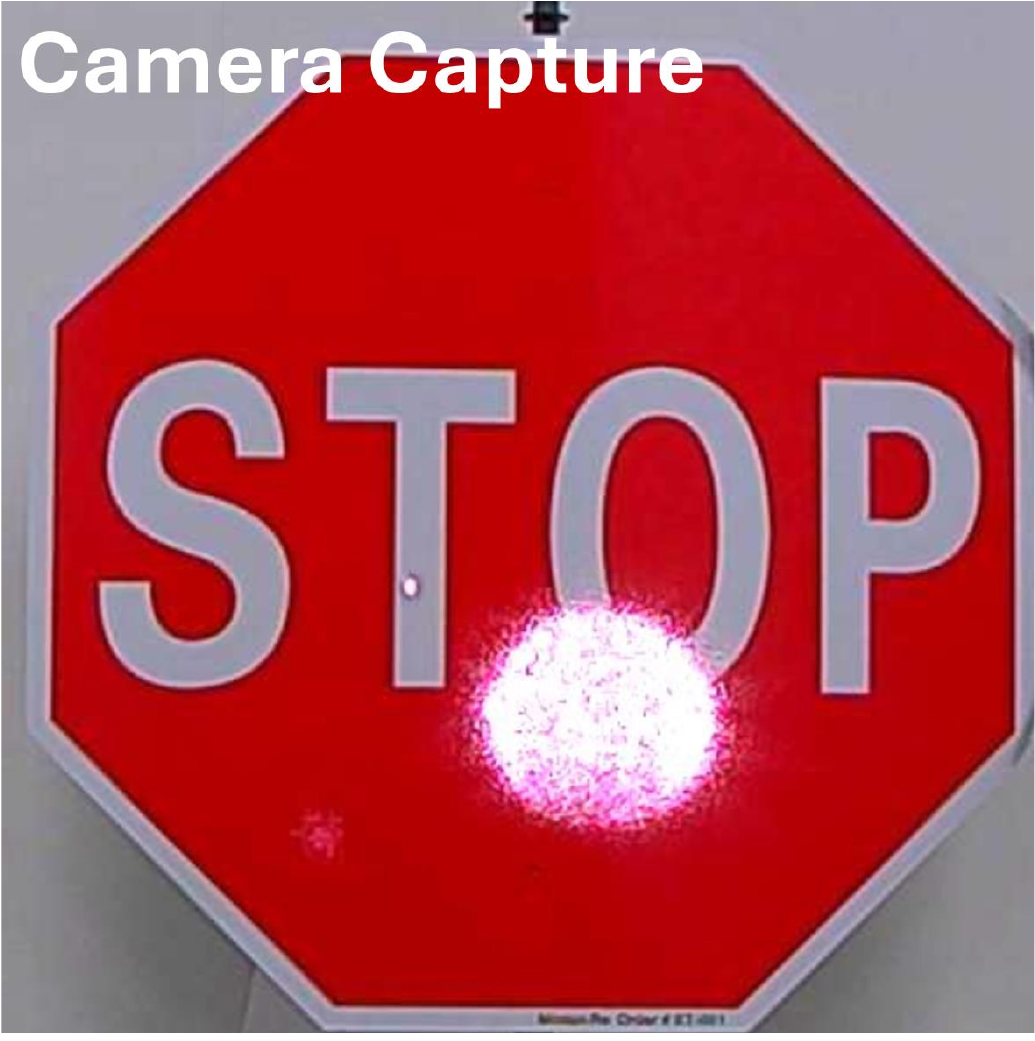}
\label{fig:patch1}
\end{minipage}
}
\subfigure[Laser]{
\begin{minipage}[t]{0.21\linewidth}
\centering
\includegraphics[width=1\textwidth]{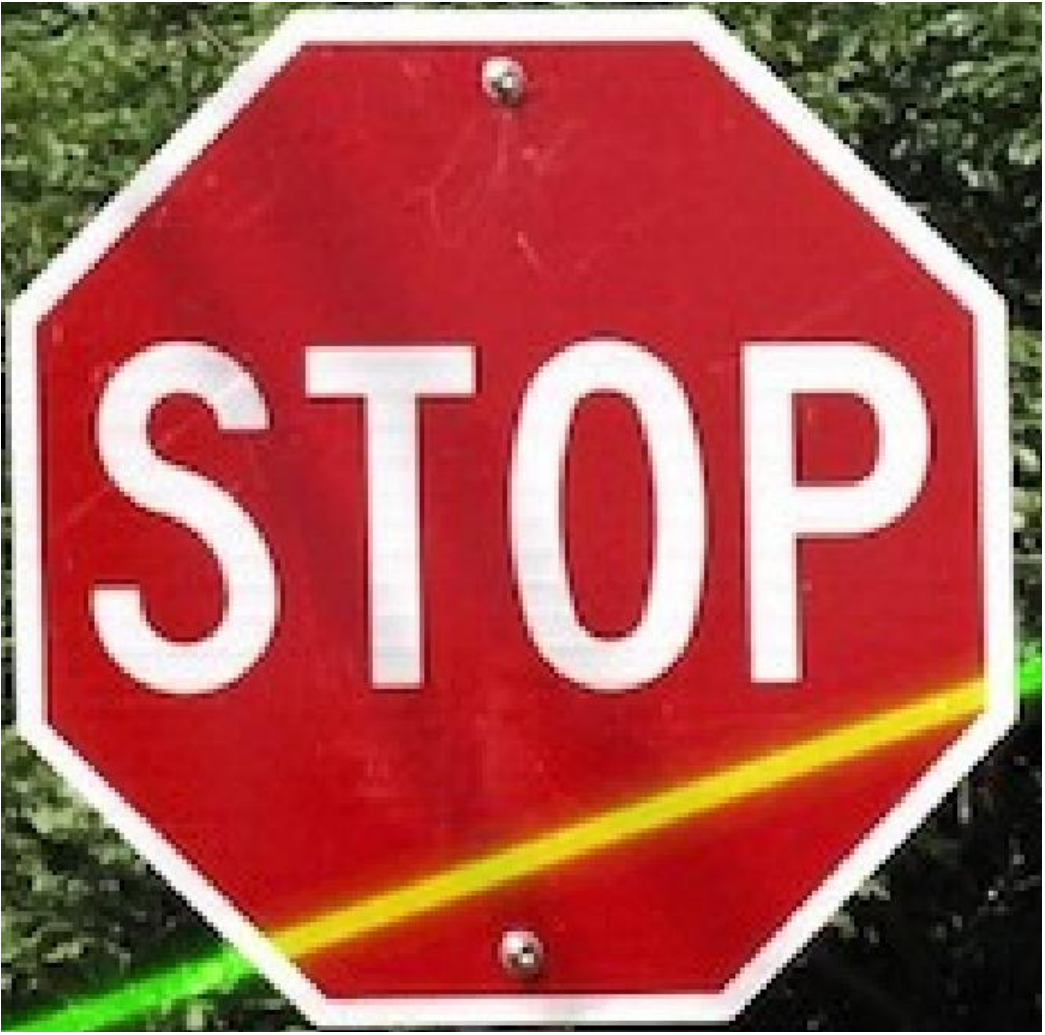}
\label{fig:patch2}
\end{minipage}
}
\subfigure[Natural light]{
\begin{minipage}[t]{0.21\linewidth}
\centering
\includegraphics[width=1\textwidth]{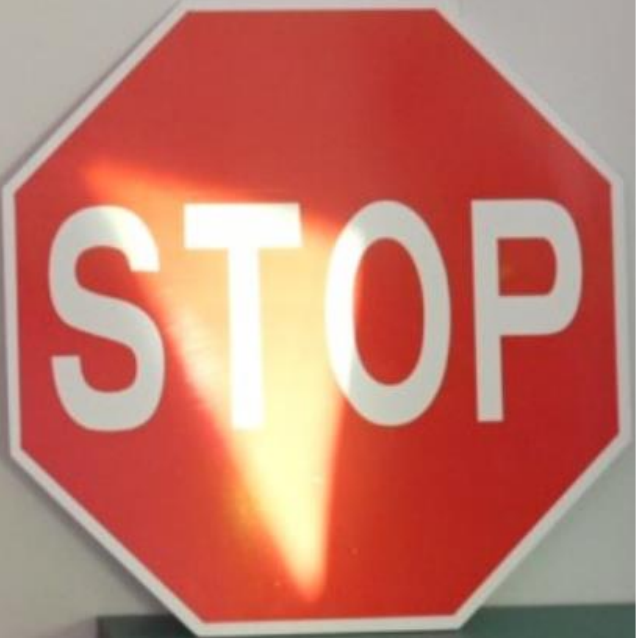}
\label{fig:patch3}
\end{minipage}
}
\subfigure[Projected light]{
\begin{minipage}[t]{0.21\linewidth}
\centering
\includegraphics[width=1\textwidth]{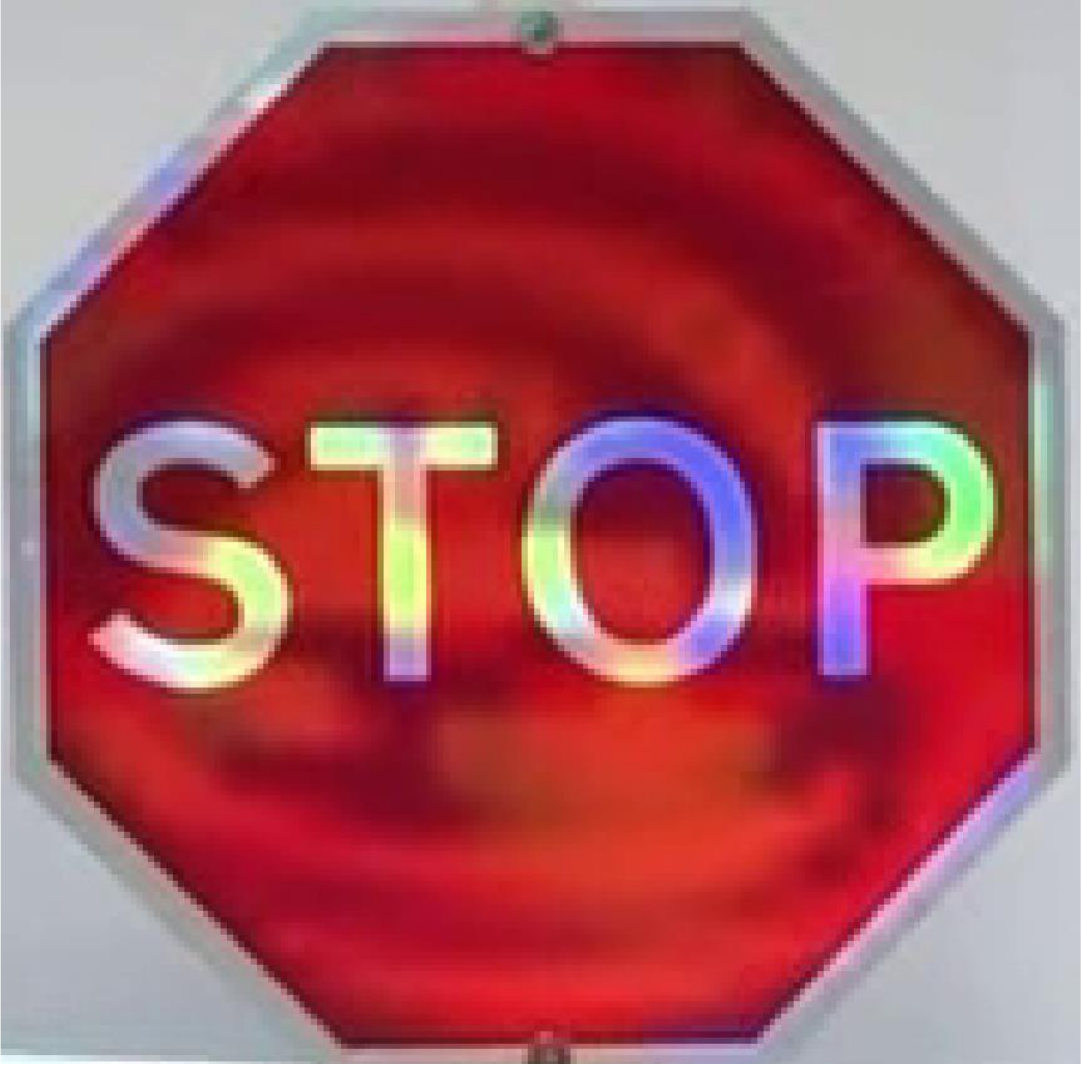}
\label{fig:patch4}
\end{minipage}
}
\caption{Four common adversarial attack modes launched by (a) infrared spot~\cite{sato2024invisible}, (b) laser line~\cite{duan2021adversarial}, (c) artificial shadow~\cite{zhong2022shadows}, and (d) projection graffiti~\cite{lovisotto2021slap}.}
\label{fig:commonPatches}
\end{figure}

The attack goal is to design and project light patches (acting as perturbation noise~\cite{zhang2021adversarial,cao2024security,wu2024clad}, making target models predict incorrect labels) onto traffic signs, and thereby disrupting original sign patterns and misleading TSRs to output false results. Light patches contain invisible and visible types according to light source properties. As shown in Fig.~\ref{fig:patch1}, the first one utilizes adversarial patches carried by infrared lights that are undetectable by human eyes but can be captured by cameras. An attacker projects an infrared spot onto the stop sign, causing TSRs to recognize it as incorrect labels. In addition, visible patches can be designed as various types as depicted in the last three subfigures of Fig.~\ref{fig:commonPatches}. Although light patches\footnote{The four light patches are designed by existing works~\cite{sato2024invisible, duan2021adversarial, zhong2022shadows,lovisotto2021slap} and we present their patterns here to facilitate understanding of the specific form of adversarial light patches.} appear in different forms, they own the same goal represented as follows: 
\begin{equation}
\label{eqn:goalFun1}
\min \psi (x + \delta ,x)  \ s.t.~{f_\theta}(x + \delta ) \ne y,
\end{equation}
where $x$ is an authentic sign image and $\delta$ is the corresponding adversarial light patch (equal to perturbation noise). $\psi(\cdot)$ denotes a distance function to measure the image variation. $f_\theta(\cdot)$ is a sign recognition model of TSR and $y$ is the true label of $x$. To solve the above constrained optimization problem, Eq.~(\ref{eqn:goalFun1}) is transformed as the Lagrangian-relaxed form~\cite{eykholt2018robust}:
\begin{equation}
\label{eqn:goalFun2}
\mathop {\arg \max }\limits_\delta \mathcal{L} ({f_\theta }(x + \delta ),y) - \lambda {\left\| \delta  \right\|_p}.
\end{equation}
where $\mathcal{L}(\cdot)$ is the cross-entropy loss function, which measures the prediction result difference between contaminated images and authentic ones. $\lambda$ is a hyper parameter to regularize the sign pattern variation caused by light patches, and $\left\| \cdot  \right\|_p$ denotes the $p$-norm. To improve the probability of misleading recognition models, the attacker attempts to maximize the loss in Eq.~(\ref{eqn:goalFun2}).

\subsection{Why Is \name Designed as a Preset Module?}
Existing defense mechanisms against light patch attacks are implemented in three distinct phases: i) Defenders erect physical barriers around signs to reduce the likelihood of malicious light sources projecting on them~\cite{zhang2023security}, ii) Hardware modifications, such as installing filters on cameras, are implemented to effectively remove malicious light components~\cite{JiaLZL0022}, and iii) Upon capturing contaminated sign examples, defenders retrain the recognition model to bolster the robustness of TSRs against adversarial patches~\cite{lovisotto2021slap}. However, as discussed in Section~\ref{sec:intro}, these approaches reveal limitations in universality. \textit{To overcome this issue, we develop a protection strategy to repair contaminated sign images before feeding them into recognition models, functioning as a preset module that works beyond the aforementioned three stages. This strategy makes \name an algorithm-level plug-and-play solution providing a universal defense against various potential patches. Based on this design concept, \name avoids the need for additional expenditures on building physical barriers, retraining models, and changing TSR hardware structures.}

\begin{figure}[b]
\centering
\includegraphics[width=0.5\textwidth]{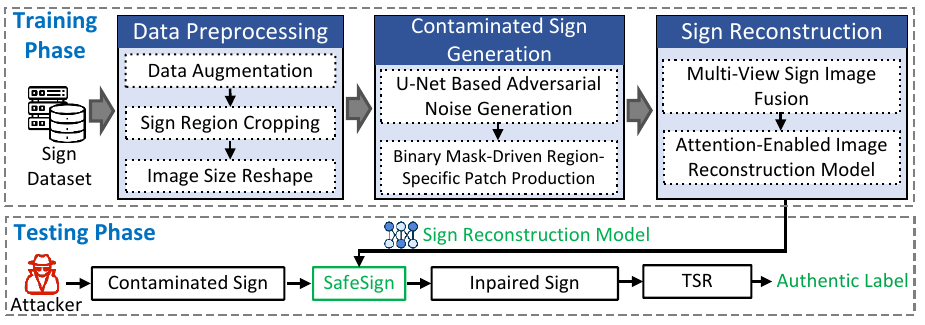}
\caption{The workflow of \sname, consisting of three parts that are data proprecessing, contaminated sign  generation, and sign reconstruction.}
\label{fig:workflowSafeSign}
\end{figure}
\section{System Design}
\label{sec:sysDesign}

\subsection{System Overview}
Fig.~\ref{fig:workflowSafeSign} depicts the fundamental workflow and corresponding technical modules of \sname. During the training phase, our primary goal is to decipher the intrinsic connection between contaminated and authentic signs, further developing an image reconstruction model that capitalizes on this relationship to repair contaminated images. In the \textit{data preprocessing} module, we first expand training data size by image augmentation operations such as affine transformation and color adjustment, which simulate variations that appears in the data collection environment and setting. Since signs always occupy partial regions (i.e., ROIs) of an image, we crop them and reshape their sizes to facilitate further processing. In the \textit{contaminated sign generation} module, we build a neural network based on U-Net architecture to introduce perturbation noise to the target sign, equivalent to imposing the impact of projecting adversarial light patches. Subsequently, binary masks are applied to control the contaminated regions, which help generate a range of potential global and local contaminated patterns. Following this, our sign image reconstruction model employs an attention mechanism to fuse complementary spatio-temporal information from multi-view images for repairing the contaminated signs. During the testing phase, \name serves as a plug-and-play preset module. Contaminated signs are first repaired by it and then fed into the TSR model for recognition. When the sign is repaired, TSRs can predict its authentic label.

\subsection{Data Preprocessing}
\label{subsec:dataAug}
\begin{figure}[t]
\centering
\subfigure[]{
\begin{minipage}[t]{0.12\linewidth}
\centering
\includegraphics[width=1\textwidth]{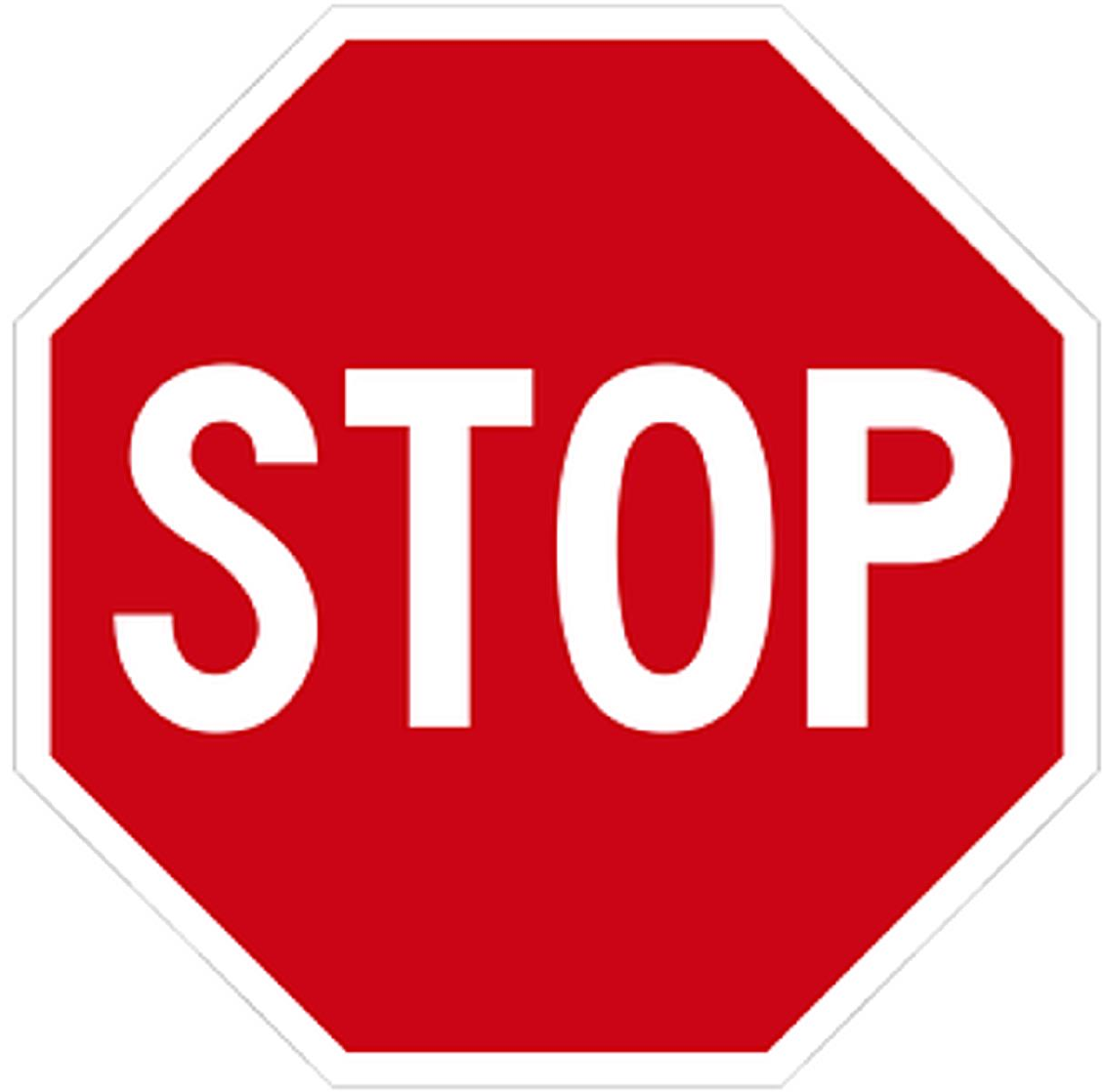}
\label{fig:augOri}
\end{minipage}
}
\subfigure[]{
\begin{minipage}[t]{0.12\linewidth}
\centering
\includegraphics[width=1\textwidth]{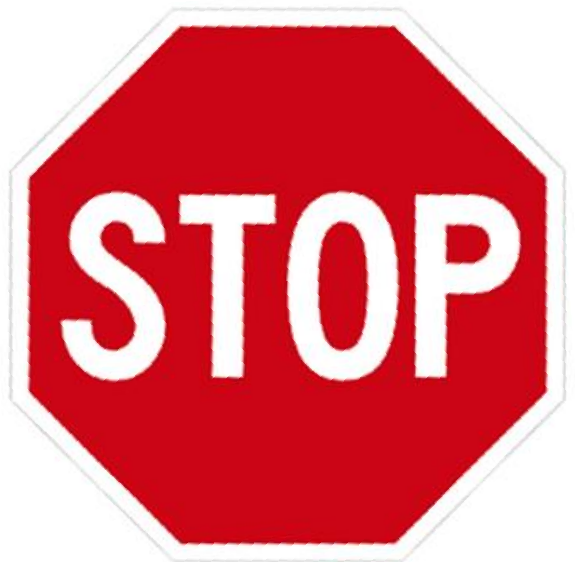}
\label{fig:aug1}
\end{minipage}
}
\subfigure[]{
\begin{minipage}[t]{0.12\linewidth}
\centering
\includegraphics[width=1\textwidth]{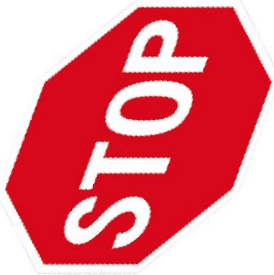}
\label{fig:aug2}
\end{minipage}
}
\subfigure[]{
\begin{minipage}[t]{0.12\linewidth}
\centering
\includegraphics[width=1\textwidth]{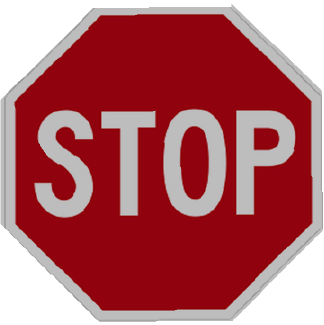}
\label{fig:aug3}
\end{minipage}
}
\subfigure[]{
\begin{minipage}[t]{0.12\linewidth}
\centering
\includegraphics[width=1\textwidth]{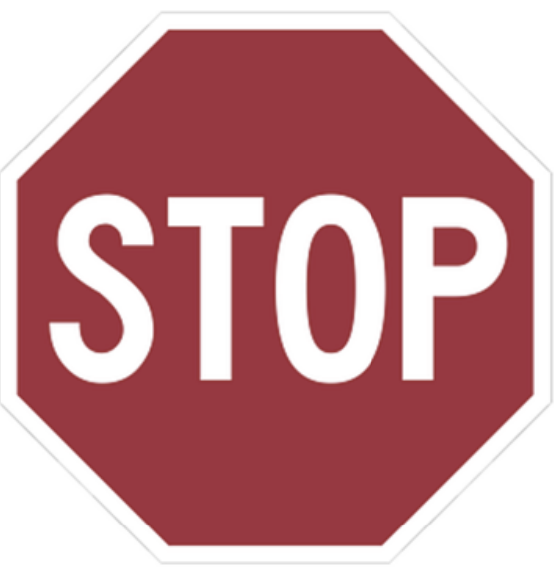}
\label{fig:aug4}
\end{minipage}
}
\subfigure[]{
\begin{minipage}[t]{0.12\linewidth}
\centering
\includegraphics[width=1\textwidth]{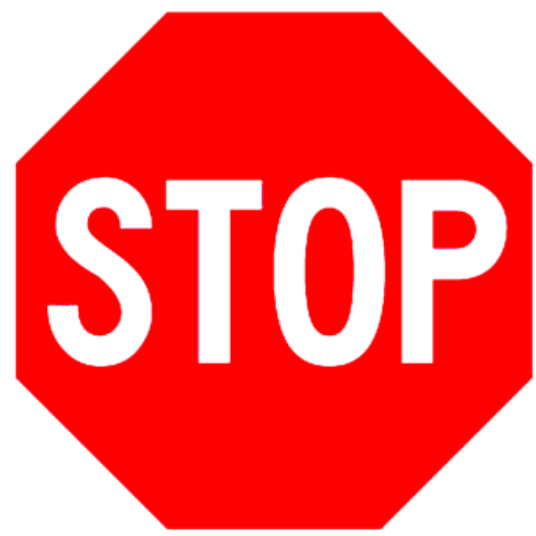}
\label{fig:aug5}
\end{minipage}
}
\caption{Sign image augmentation using five common transformation operations on (a) a raw sign image, that are adjusting (b) rotation, (c) shearing, (d) brightness, (e) saturation, and (f) contrast.}
\label{fig:augData}
\end{figure}

\textbf{Data augmentation.}
Providing ample training data is helpful for recognition models in comprehending the structure and embedded information of sign images. Nonetheless, collecting sign images in real-world environments is a non-trivial task, as it demands significant time and labor costs. Therefore, we design a data augmentation method by conducting image transformation operations in the digital world. It considers the variation of collected sign images when vehicles move, caused by changes in light condition and relative shooting position. We rely on two changes to augment the data volume of sign images. Firstly, as the spatial relationship between a vehicle and target signs varies, so does the ROI size and shape within the captured image. Such variations can be precisely modeled using image affine transformations~\cite{dong2002affine,3411806}. In geometric terms, this transformation constitutes a linear transformation of a vector space (i.e., multiplying by matrix \textit{A}) and is coupled with the addition of a translation vector $\boldsymbol{c}$, thereby achieving a space transformation. This mathematical manipulation allows for the execution of various image transformations, such as angle rotation and shape shearing. Given the coordinate $\left({b_0,b_1}\right)$ of a pixel in the original sign image, after the affine transformation, the new coordinate of this point is changed to $\boldsymbol{d}$:
\begin{equation}
\label{eqn:affineT1}
\boldsymbol{d} = \boldsymbol{A}\left[ {\begin{array}{*{20}{c}}
b_0\\
b_1
\end{array}} \right] + \boldsymbol{c}
\end{equation}
To better understand the matrix structure, Eq.~(\ref{eqn:affineT1}) can be expanded as follows:
\begin{equation}
\label{eqn:affineT2}
\left[ {\begin{array}{*{20}{c}}
d_0\\
d_1
\end{array}} \right] = \left[ {\begin{array}{*{20}{c}}
{{a_{00}}}&{{a_{01}}}\\
{{a_{10}}}&{{a_{11}}}
\end{array}} \right] \cdot \left[ {\begin{array}{*{20}{c}}
b_0\\
b_1
\end{array}} \right] + \left[ {\begin{array}{*{20}{c}}
{{c_{0}}}\\
{{c_{1}}}
\end{array}} \right]
\end{equation}
We further express this kind of affine transformation in a single matrix, by utilizing the homogeneous coordinate matrix to reform the above form. 
\begin{equation}
\label{eqn:affineT3}
\left[ {\begin{array}{*{20}{c}}
d_0\\
d_1\\
1
\end{array}} \right] = \left[ {\begin{array}{*{20}{c}}
{{a_{00}}}&{{a_{01}}}&{{c_0}}\\
{{a_{10}}}&{{a_{11}}}&{{c_1}}\\
0&0&1
\end{array}} \right] \cdot \left[ {\begin{array}{*{20}{c}}
b_0\\
b_1\\
1
\end{array}} \right]
\end{equation}
By adjusting the parameters of $\boldsymbol{A}$ and $\boldsymbol{c}$, we can perform the necessary transformation operation on original sign images, effectively expanding the dataset. On the other hand, lighting condition variations prompt adjustments in shooting parameters such as shutter speed and aperture, consequently modifying brightness, saturation, and contrast of captured sign images. Brightness denotes the overall luminosity, while saturation signifies the purity and vibrancy of colors in one image. Contrast pertains to the variance in brightness or color among different image subregions. These factors collectively affect the visual impact and image quality. In our work, \name manipulates them to produce multiple derivative versions of raw sign images. Fig.~\ref{fig:augData} presents five common augmentation versions of a ``STOP" sign image. We observe that the derivative version retains the traffic sign information, but makes adjustments in size, shape, and color.

\textbf{Sign Region Cropping \& Size Reshape.}
Regardless of the architecture type of TSRs, the contamination target of light patch-based attacks is the traffic sign region. After capturing an image that includes the surrounding scene, the first step is to employ a traffic sign detector to identify ROIs. For both single-stage and two-stage recognition models, once these ROIs are detected, we add perturbation noise to these regions. Following this, the contaminated images are forwarded to the recognition model. To address the size inconsistency of inputs, we standardize them to facilitate further processing. In our study, the dimensions of all inputs are normalized to $64 \times 64$.

\subsection{Contaminated Sign Generation}
\label{sec:noiseGen}
\name aspires to develop an image inpainting mechanism designed to eliminate the disruptive effects of light patches on traffic signs. This innovation can ensure the performance of sign recognition models when processing adversarial examples and bolster their robustness against potential attacks. This repair capability depends on a thorough understanding of the mapping relationship between authentic and contaminated sign pairs. To achieve this goal, the first step is to collect contaminated signs for training the model. An intuitive approach to obtain them involves reproducing light patch-based attacks and gathering corresponding sign images. However, this approach requires significant resource investment and may interrupt normal traffic services, making it impractical. To overcome this issue, we analyze the attack launching process and reveal its fundamental principle: regardless of attack modes and patch patterns, they all aim to disrupt original feature patterns in specific regions of authentic signs, thereby leading to the misjudgment of TSRs. Therefore, our task for collecting contaminated signs naturally evolves into introducing adversarial perturbation noise to authentic sign images. The noise's impact mirrors that of projecting malicious light patches, as both strive to disrupt original sign patterns.

\begin{figure}[t]
\centering
\includegraphics[width=0.47\textwidth]{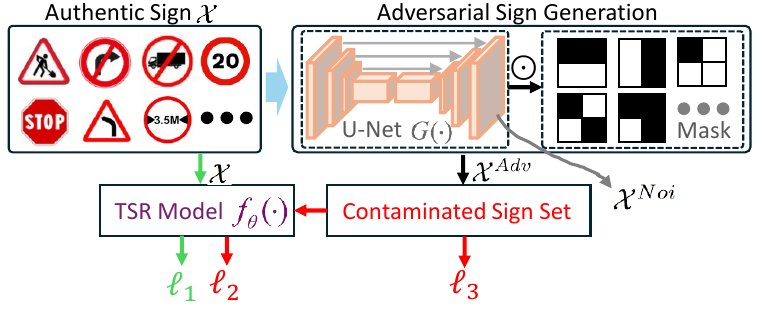}
\caption{Architecture of contaminated sign generation module: an U-Net based neural network utilized to generate raw perturbation noise and binary masks in charge of controlling the contaminated region.}
\label{fig:advGen}
\end{figure}

Based on this insight, we propose a binary mask-driven U-Net module to generate diverse perturbation noise patterns and corresponding contaminated signs. The reason for using the U-Net framework is its encoder-decoder structure, which effectively captures both global and local image information, excelling at common image processing tasks. The overall architecture of our adversarial sign generation module is depicted in Fig.~\ref{fig:advGen}. We denote authentic signs as $\mathcal{X} = \{ {X_1},{X_2},..., {X_m}, ..., {X_M}\}$, where ${X_m} = \{ x_{m,1},x_{m,2},...,x_{m,n},..., x_{m,N}\}$ is the image set of $m$-th sign type and $N$ is the total image number of this type. $\mathcal{X}$ is first fed into the U-Net neural network, which outputs perturbation noise $\mathcal{X}^{Noi}$ with the same size as the input. This raw noise pattern is then subjected to element-wise multiplication with a binary mask $\mathcal{B}$ to control its contaminated regions. Following this, region-specific noise is added to the original input image to generate adversarial samples (i.e., the contaminated signs), described as $\mathcal{X}^{Adv} = \mathcal{X} + \mathcal{X}^{Noi} \odot \mathcal{B}$. Finally, these contaminated signs are input into a trained sign recognition model $f_\theta(\cdot)$, where $\odot$ is Hadamard product. Our contaminated sign generation module aims to maximize the misrecognition rate and this process is described as follows: 
\begin{equation}
\label{eqn:affineT3}
\mathop {\max }\mathcal{L}({f_\theta }(\underbrace {{x_{m,n}} + G({x_{m,n}}) \odot \mathcal{B}}_{x^{Adv}_{m,n}}),{y_{m,n}}),
\end{equation}
where $y_{m,n}$ is the true label of sign image $x_{m,n}$ and $x^{Adv}_{m,n}$ is a contaminated version of $x_{m,n}$. Among them, the recognition model ${f_\theta}(\cdot)$ is trained by minimizing the cross-entropy loss (namely, $\ell_1$) between authentic and predicted sign labels to enhance the recognition accuracy. For the U-Net based network, we optimize its parameters by maximizing the loss (namely, $\ell_2$) described in Eq.~(\ref{eqn:affineT3}). However, we observe that the above optimization design only encourages $G(\cdot)$ to generate contaminated signs that effectively fool recognition models but cannot ensure the internal diversity of contaminated patterns. That is, multiple adversarial noise versions against one sign exhibit similar patterns, which is inconsistent with our goal of generating various authentic and contaminated image pairs. By examining the sign generation model architecture, we reveal the reason for this issue is that the chosen loss function tends to cause the optimization to fall into local minima. In this case, $G(\cdot)$ cannot thoroughly learn the potential pattern of contaminated signs. To overcome this issue, a new loss component $\ell_3$ is introduced into the loss function to drive $G(\cdot)$ to fully explore the feature space of perturbation noise, thus producing diverse contaminated signs. $\ell_3$ is given as follows:
\begin{equation}
\label{eqn:affineT4}
\max \sum\limits_{{p_1} = 1}^P {\sum\limits_{{p_2} = 1}^P {\xi (x_{m,n}^{Noi,{p_1}},x_{m,n}^{Noi,p_2})}},
\end{equation}
where $P$ is the total number of perturbation noise patterns of $x^{Noi}_{m,n}$ and $\xi(\cdot)$ is the mean absolute error to measure the differences among these examples. The index $p_1$ is not equal to $p_2$. Maximizing $\ell_3$ directly drives the generated noise patterns to distribute in different positions of high-dimensional feature embedding space, thus enhancing the pattern diversity of generated samples. Since the sign recognition model (related to $\ell_1$) is trained before, we hereby only need to consider how to optimize the adversarial sign generation model. Therefore, the final loss function consists of two components, which deceives the TSR model while ensuring sample diversity, given by:
\begin{equation}
\label{eqn:affineT5}
L_{total} = \alpha \ell_2 + \beta \ell_3.
\end{equation}
where $\alpha$ and $\beta$ are both scalar weights in the range of $[0,1]$. Given the equal importance of deceiving the recognition model and preserving the diversity of contaminated signs, we have set both weights to 0.5 for our study in this paper.
\begin{figure}[b]
\centering
\subfigure[One horizontal or vertical segmentation line]{
\begin{minipage}[t]{1\linewidth}
\centering
\includegraphics[width=0.9\textwidth]{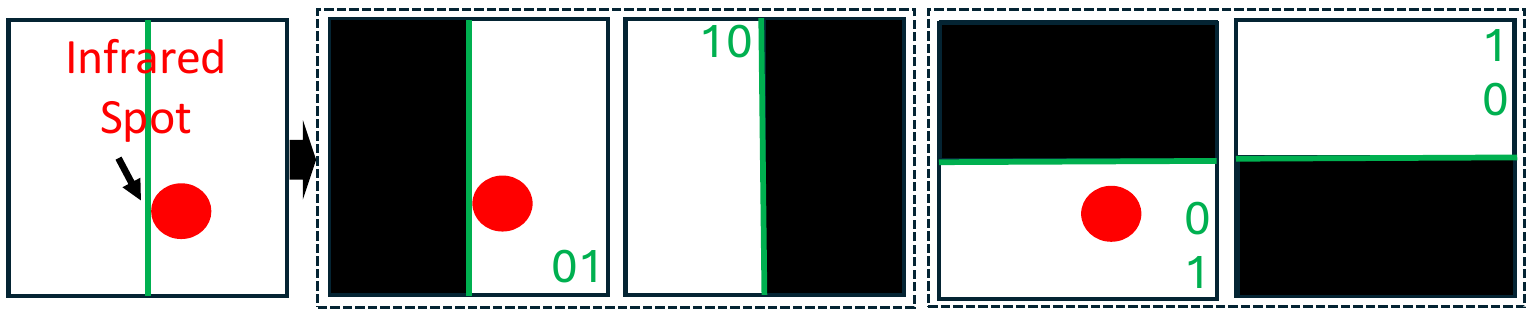}
\label{fig:mask1}
\end{minipage}
}
\subfigure[Two or six segmentation lines]{
\begin{minipage}[t]{1\linewidth}
\centering
\includegraphics[width=0.89\textwidth]{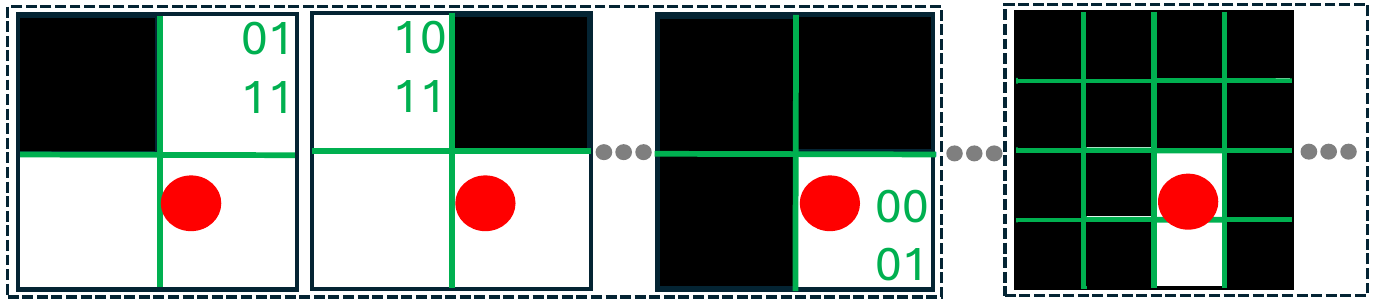}
\label{fig:mask2}
\end{minipage}
}
\caption{Mask patterns corresponding to different contaminated blocks when using (a) one, (b) two and six segmentation lines, respectively.}
\label{fig:maskpattern}
\end{figure}

Moreover, the binary mask plays a critical role in \sname, which determines the regions where perturbation noise is added. We hereby introduce how to design the mask generation strategy that adequately covers potential existing regions of light patches. To achieve this, we first explore the working principles of binary masks. As illustrated in Fig.~\ref{fig:maskpattern}, each mask matches the input image in size and is divided into multiple equal-sized blocks, labeled as either 0 or 1, representing black and white regions, respectively. The blocks labeled 1 are designated regions for adding perturbation noise, while the black blocks are not. By multiplying the original perturbation noise $\mathcal{X}^{Noi}$ with the mask $\mathcal{B}$ and adding their products to $\mathcal{X}$, the contaminated sign is generated. Subsequently, we utilize the infrared spot depicted in Fig.~\ref{fig:patch1} to introduce how to design the mask to assist in generating perturbation noise. The mask divides the sign image using horizontal and vertical green segmentation lines, with block size inversely proportional to the number of lines. As shown in Fig.~\ref{fig:mask1}, one horizontal or one vertical line divides the image into two parts and generates four mask modes. We can observe that the noise adding blocks of two cases (i.e., $\left[ {01} \right]$ and $\footnotesize \left[ \begin{array}{l}0\\1\end{array} \right]$) covers the patch existing region. By increasing the segmentation lines, the region size of each block correspondingly decreases and the search granularity of light patches gradually enhances. For example, as shown in Fig.~\ref{fig:mask2}, by simultaneously utilizing two lines (i.e., one horizontal and one vertical), we can search the patch region within one-fourth of the image. In contrast, with six lines, we can narrow the search region to one-sixteenth of the image. This method allows for flexible adjustment of the size and granularity of noise-added regions by modifying the number of segmentation lines. While we hereby leverage the case of light patches in a single region to explain the mask design, our method is equally applicable to patches distributed across multiple regions.
\begin{figure}[t]
\centering
\subfigure[Four contaminated patterns]{
\begin{minipage}[t]{0.42\linewidth}
\centering
\includegraphics[width=1\textwidth]{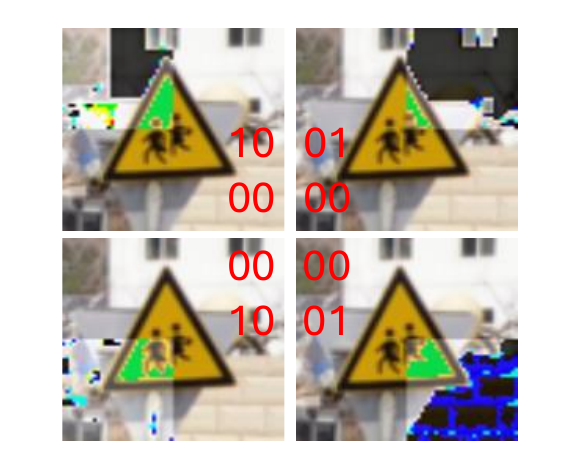}
\label{fig:noisePattern}
\end{minipage}
}
\subfigure[Sign recognition accuracy]{
\begin{minipage}[t]{0.46\linewidth}
\centering
\includegraphics[width=1\textwidth]{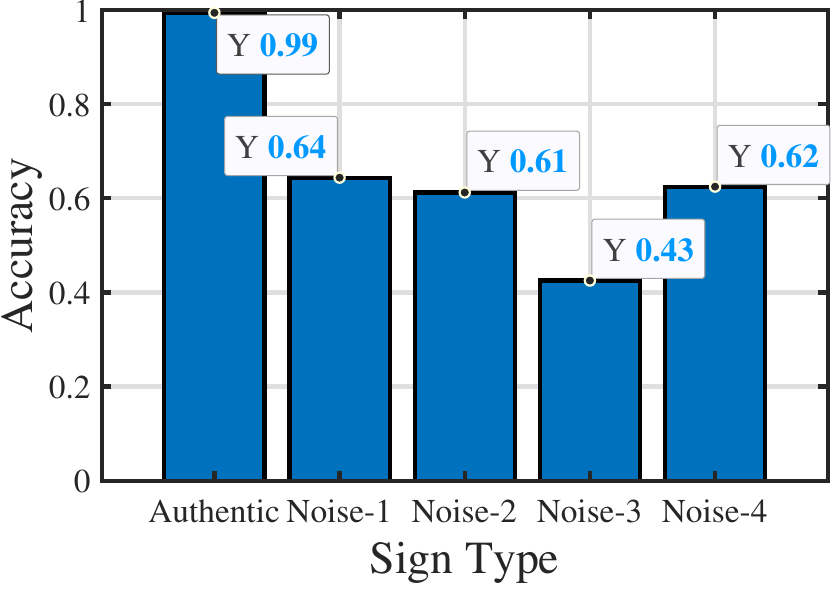}
\label{fig:noiseAccuracy}
\end{minipage}
}
\caption{Effectiveness of our contaminated sign generation module: four perturbation noise patterns imposed on different sub-regions shown in (a) and corresponding sign recognition performance depicted in (b).}
\label{fig:maskAcc}
\end{figure}

\begin{figure*}[b]
\centering
\includegraphics[width=0.92\textwidth]{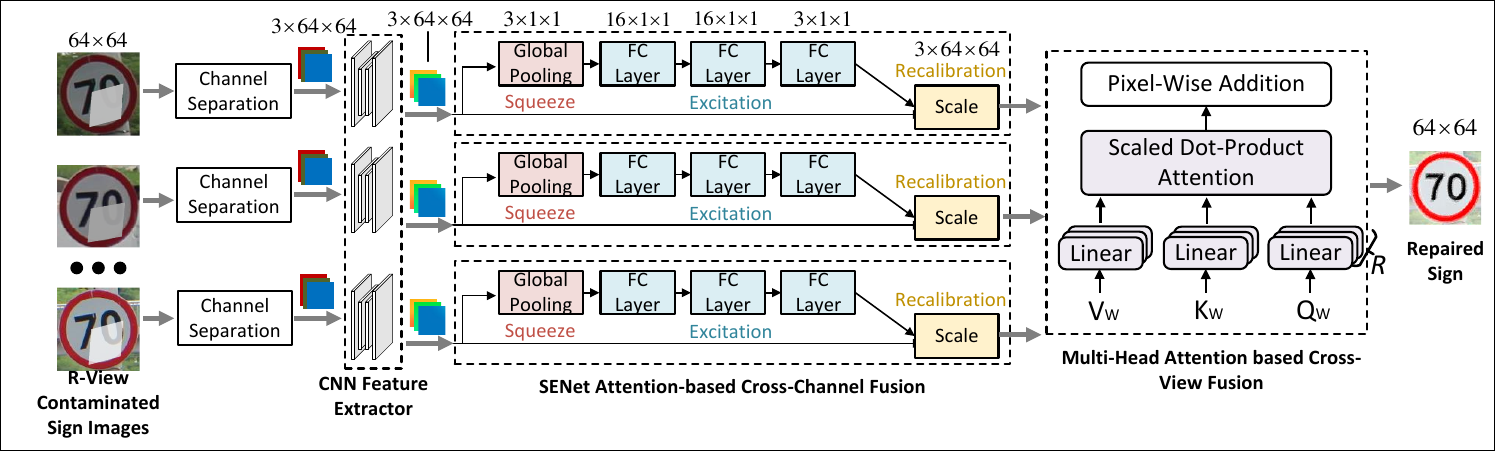}
\caption{The framework of the attention mechanism-empowered sign reconstruction model: jointly using the information of multi-view contaminated images to remove the impact of infrared light patches.}
\label{fig:recFramework}
\end{figure*}

Subsequently, an experiment provides a preliminary understanding of the impact of perturbation noise on TSR performance. We first utilize the traffic sign dataset~\cite{data1} containing 6164 traffic sign images from 58 classes, to train our contaminated sign generation model. $G(\cdot)$ adds perturbation noise to four sub-regions corresponding to using mask patterns as $\footnotesize\left[ {\begin{array}{*{20}{c}} 1&0\\0&0 \end{array}} \right]$, $\footnotesize\left[ {\begin{array}{*{20}{c}} 0&1\\0&0 \end{array}} \right]$,
$\footnotesize\left[ {\begin{array}{*{20}{c}} 0&0\\1&0 \end{array}} \right]$, and $\footnotesize\left[ {\begin{array}{*{20}{c}} 0&0\\0&1 \end{array}} \right]$. These adversarial examples are then input into the trained recognition model and we count the classification accuracy variation under without and with adding noise cases. To understand the specific form of our perturbation noise, we display four contaminated examples against the sign ``Students on the Road" in Fig.~\ref{fig:noisePattern}. Additionally, Fig.~\ref{fig:noiseAccuracy} displays the accuracy variation as imposing the perturbation noise to disrupt original sign patterns, with an average decline of 41.5\% compared to the original performance (i.e., 98.5\%). The results highlight that the generated adversarial signs can significantly deceive recognition models, and present the effectiveness of our contaminated sign generation module.

\subsection{Sign Image Reconstruction}
\label{sec:imagerec}
This section introduces how to develop an attention-powered reconstruction model to fuse multi-view images for repairing contaminated signs. This model design is inspired by two key insights. First, onboard cameras continuously record images from different views while in motion, each subjected to different lighting conditions and shooting settings, thus capturing correlated spatial and temporal information about the same contaminated traffic sign. In this context, the impact of light patches on sign patterns differ across these multi-view images. Consequently, compared with single-image inputs, multi-view data provide more spatial-temporal information that is beneficial for image repair. Second, since each image gives unique contributions to the sign repair process, it is crucial to effectively fuse and utilize them. Considering the power of attention mechanisms in information fusion~\cite{liu2022attention,lientionfgan}, we harness this capability to leverage multi-view images for completing image repair. We detail the technical framework of our reconstruction model in Fig.~\ref{fig:recFramework}, which consists of three main components: a CNN-based feature extractor, a SENet-based cross-channel attention module, and a cross-view self-attention module. Each input fed into this model includes $R$ images from different views and each image contains three color channels: red, green, and blue. We denote the input as $\mathcal{X}_{m}^{In} = \left\{ {x_{m,1}^{Adv,red/green/blue},...,x_{m,R}^{Adv,red/green/blue}} \right\}$. The $R$ images of $\mathcal{X}_{m}^{In}$ are randomly selected from the $m$-th sign type.
The input images are initially processed by $R$ parallel feature extractors, each consisting of three standard CNN units. CNNs are utilized for their exceptional performance to eliminate redundant information and extract representative non-linear features~\cite{cao2022handkey,3613288,3623153}. Subsequently, the extracted feature maps are fed into SENet networks to complete content-aware and weighted information fusion across three color channels of each image. The cross-channel attention operation is based on the understanding that traffic signs exhibit diverse color distributions, thus each channel distinctively contributes to various traffic sign patterns. The SENet module calculates the information weight (i.e., the contribution for sign repair) of each channel through three steps: squeezing, excitation, and recalibration. Squeezing operation compresses the spatial dimension information into a channel descriptor using a global average pooling layer, which helps highlight the global importance of each channel. Excitation uses a fully connected layer to learn the dependencies between channels and applies an activation function to obtain the importance weights for each channel. Recalibration multiplies the resultant weights with the original feature maps to achieve weighted information fusion across the channels. For cross-image fusion, we develop a standard multi-head self-attention module~\cite{zhao2024ssir} to capture the information interaction among multi-view images. It projects $R$ images of each input sequence into $R$ different weight matrices (i.e., $V^{1,2,...,R}_w$, $K^{1,2,...,R}_w$, and $Q^{1,2,...,R}_w$), then computes them in parallel and merges the results from each group as the final repaired sign by scaled dot-product attention and pixel-wise addition~\cite{yu2022paramixer}. By using the multi-head attention mechanism, the proposed framework can better capture multi-view features from the input, enhancing model generalization and reducing the risk of overfitting. In a nutshell, our reconstruction model outputs a repaired image that matches the input size, and its training goal is formalized as follows:
\begin{equation}
\label{eqn:affineT9}
\mathop {\arg \min }\limits_S \mathcal{L}({f_\theta}(S(\mathcal{X}_{m}^{in})),y_m)
\end{equation}
where $S(\cdot)$ is the reconstruction model and $y_m$ is the authentic label. To optimize the parameters of $S(\cdot)$ and improve the sign repair performance, we minimize the cross-entropy loss of ${f_\theta }( \cdot )$. Moreover, the training data of our reconstruction model consists of authentic sign images, to ensure that $S(\cdot)$ can guarantee its applicability with authentic inputs, which has been minutely introduced in Section~\ref{subsec:expSetting}.

\begin{figure}[t]
\centering
\subfigure[Feature map similarity distribution of three-type sign images]{
\begin{minipage}[t]{0.46\linewidth}
\centering
\includegraphics[width=1\textwidth]{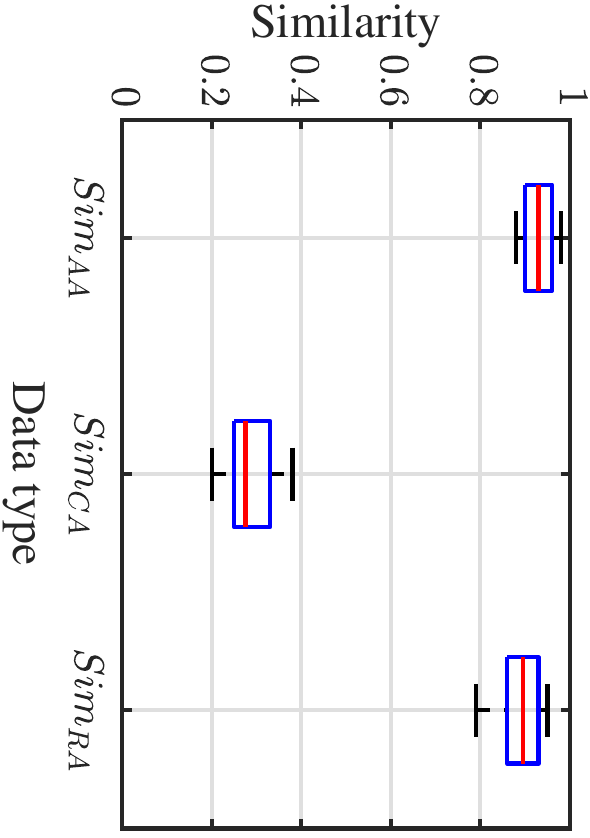}
\label{fig:repairFeature}
\end{minipage}
}
\subfigure[Recognition accuracy after repairing four-type contaminated signs]{
\begin{minipage}[t]{0.46\linewidth}
\centering
\includegraphics[width=1\textwidth]{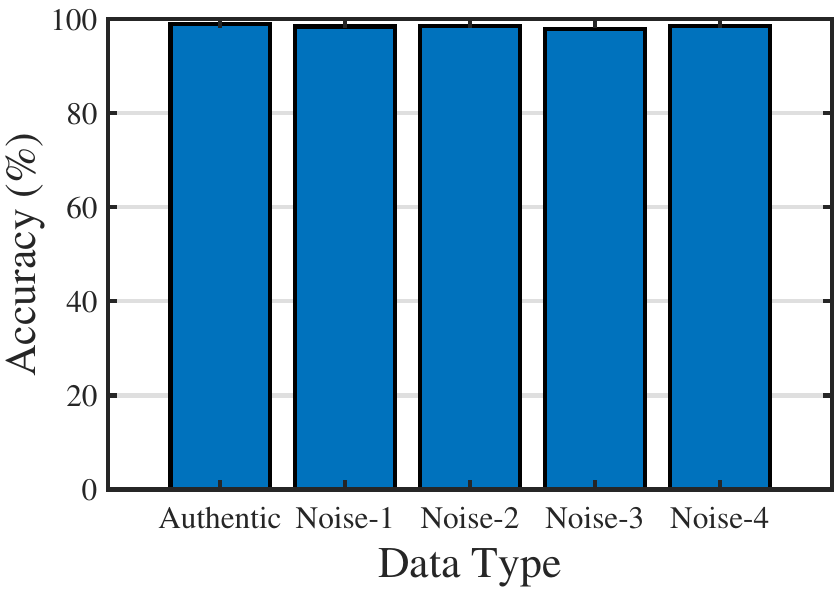}
\label{fig:repairAcc}
\end{minipage}
}
\caption{Repair ability evaluation of our reconstruction model by measuring the feature map difference in (a) and the recognition performance as inputting repaired sign images in (b).}
\label{fig:repairPerformance}
\end{figure}

To explore the repair ability of our reconstruction model, we input the contaminated sign images (corresponding to four masks as described in Section~\ref{sec:noiseGen}) into $S(\cdot)$ and obtain their repaired versions. Subsequently, we calculate the feature map differences (measured by Cosine similarity) outputted by all network layers when inputting three types of samples (i.e., authentic, contaminated, and repaired images) into $f_\theta(\cdot)$. As illustrated in Fig.~\ref{fig:repairFeature}, the feature similarity of the repaired images to the authentic ones (namely, $Sim_{RA}$) is close to the internal similarity of authentic images (denoted as $Sim_{AA}$), which is markedly higher than that of the contaminated images (referred to as $Sim_{CA}$). Moreover, the recognition accuracy of the repaired images depicted in Fig.~\ref{fig:repairAcc} reaches 97.2\%, which is very close to that of the authentic ones. The experiment results demonstrate that the reconstruction model effectively restores the critical information of traffic signs, ensuring that the TSR recognizes them with high accuracy.

\subsection{Put All Things Together}
We hereby revisit the principal technical components involved in the implementation process of \name and delve deep into their working mechanisms. The primary objective of \name is to repair sign images destroyed by malicious light patches, accomplished through our reconstruction model. This countermeasure ensures high accuracy of our redesigned TSR system, even when facing potential light patch attacks. To fully optimize the model parameters, the fundamental step is to provide a substantial number of paired authentic and contaminated images for training. To meet this requirement, we augment the original dataset through image transformation operations and design a contaminated sign generation model to produce sufficient contaminated data. In this way, the trained reconstruction model is well-equipped to counteract threats imposed by adversarial light patches.

\section{Evaluation}
\label{sec:evaluation}
In this section, we first introduce our experiment settings, including sign recognition models, traffic sign datasets, light patch-based attack approaches, performance evaluation metrics, and the model construction process. We then evaluate the performance of \name against light patch-based attacks.

\subsection{Experiment Settings}
\label{subsec:expSetting}

\textbf{Sign recognition models.} In this study, we examine two types of traffic sign recognition models referring to the setting of existing works~\cite{lovisotto2021slap,duan2021adversarial,zhong2022shadows,sato2024invisible}: the single-stage model YOLO5, and the two-stage models LeNet and GoogleNet. YOLO5, a widely used traffic sign detector, consists of twenty-four convolutional layers and two fully connected layers, incorporating residual blocks. It processes the entire image captured by a camera to directly predict the type and position of traffic signs. For the two-stage models, we select LeNet and GoogleNet, both known for their excellent and stable performance in various image recognition tasks. LeNet is composed of three convolutional layers, two pooling layers, and three fully connected layers. GoogleNet features a network depth of twenty-seven layers and includes inception units to enhance performance.

\textbf{Traffic sign datasets.}
Three public traffic sign datasets are used to build and evaluate TSR models: GTSRB~\cite{dataset1}, CTSD~\cite{dataset2}, and PTSD~\cite{dataset3}. These datasets include traffic sign images collected from multiple countries/areas, such as America and Europe, covering common signs utilized in daily life. GTSRB contains more than 50,000 images spanning 43 traffic sign categories, with annotations for the coordinates of each sign. CTSD includes 6,164 traffic sign images across 58 categories. PTSD comprises 43 sign types with a total of 16,000 images. As described in Section~\ref{subsec:dataAug}, to enable a model to better learn the traffic signs and optimize parameters, we develop a data augmentation module. In our study, we utilize eight types of image transformation approaches to increase the data scale to eight times its original size, providing enough training data for our sign reconstruction model. Relying on the three datasets, we generate the same number of contaminated images as authentic ones for training our reconstruction models as described in Section~\ref{sec:noiseGen}.

\textbf{Light patch-based attack approaches.}
To verify the performance of our \sname, we employ it to repair traffic signs contaminated by light patches, which have been proven to effectively deceive TSRs. Among existing adversarial light patches, we select four representative types: infrared spots (IS)~\cite{sato2024invisible}, laser lines (LL)~\cite{duan2021adversarial}, natural light shadows (NLS)~\cite{zhong2022shadows}, and projective graffiti (PG)~\cite{lovisotto2021slap}. IS utilizes long-range, human-invisible infrared lights to induce erroneous judgments by TSRs. LL employs laser beams to distort the original sign patterns for enabling attacks. NLS achieves a high degree of concealment by reflecting natural sunlight with specific shapes onto the target sign. PG involves using a projector to remotely project specially designed light graffiti. For each attack approach, we generate light patches based on their design principles. The number of four-typed contaminated sign images generated for deceiving TSRs are all three thousand. Both the contaminated and repaired signs are then fed into recognition models, and we record the corresponding recognition performance. By analyzing the accuracy variation in two cases, we can verify the effectiveness of \sname.

\textbf{Metrics.}
\label{subsec:metrics}
Traffic sign recognition is fundamentally a multi-class classification task. Therefore, we use \textit{accuracy} to measure the proportion of correctly classified signs in the entire dataset. Considering the uneven distribution of images across different sign types, we employ \textit{precision} to assess the classification accuracy for each type. A satisfactory sign repair mechanism and secure TSR system should ensure high accuracy and precision. 

\textbf{Model construction.}
\label{subsec:modelCon}
In our work, we need to train three models: sign recognition model, contaminated sign generation model, and reconstruction model. For the first one, we adhere to their original network architecture and training settings. The generation model construction relies on the U-Net framework. U-Net features a symmetric network structure, consisting of an encoder and a decoder. As described in Fig.~\ref{fig:advGen}, our encoder consists of eight consecutive convolutional units, each followed by a BatchNorm layer and a ReLU activation function. A pooling operation is performed after every two units and their output channels are 64, 32, 64, and 8, respectively. The right half uses symmetric deconvolutional layers to restore the image size and output perturbation noise. There are skip connections between the encoder and decoder, linking the output of the corresponding layers in the encoder to the input of the corresponding layers in the decoder. This design allows the original sign feature maps to be directly passed to the decoder, assisting the network in learning specific malicious patches for particular signs. Additionally, we leverage four types of masks (using 1, 2, 4, and 6 segmentation lines, respectively) to control the regions where perturbation noise is added. Our reconstruction model's CNN feature extraction module includes three standard convolutional layers, each followed by a ReLU activation function and a BatchNorm layer. SenseNet employs a standard architecture, comprising three modules, namely, squeezing, excitation, and recalibration. As shown in Fig.~\ref{fig:recFramework}, the three fully connected (FC) layers have 16, 16, and 3 neurons respectively. The first two FC layers are followed by a ReLU activation function, while the last one is followed by a Sigmoid activation function. The cross-view attention module has an input size of $R\times 3\times 64 \times 64$, using $R\times3$ standard weight matrices to complete the attention based information fusion, finally outputting a repaired sign with a size of $64\times64$. During the training phase, the parameter settings for these models are consistent. We employ the predefined training and test set ratios from the three datasets to split the data. The model parameters are optimized using the augmented data by default. The dropout rate is set to 0.5, and the batch size is 32. We use the Adam optimizer with a learning rate of $1{e^{-4}}$.

\begin{figure}[b]
\centering
\subfigure[Accuracy of YOLO5]{
\begin{minipage}[t]{0.29\linewidth}
\centering
\includegraphics[width=1\textwidth]{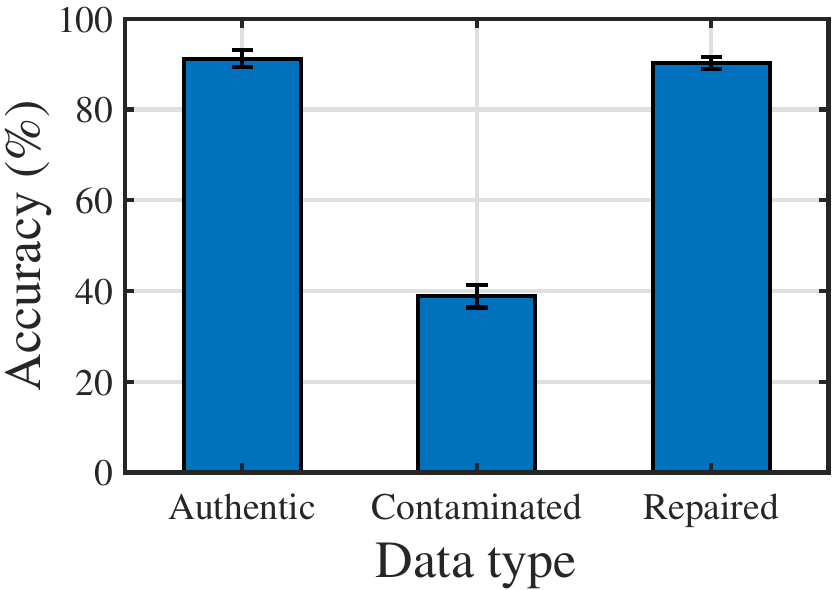}
\label{fig:overallAcc1}
\end{minipage}
}
\subfigure[Accuracy of LeNet]{
\begin{minipage}[t]{0.29\linewidth}
\centering
\includegraphics[width=1\textwidth]{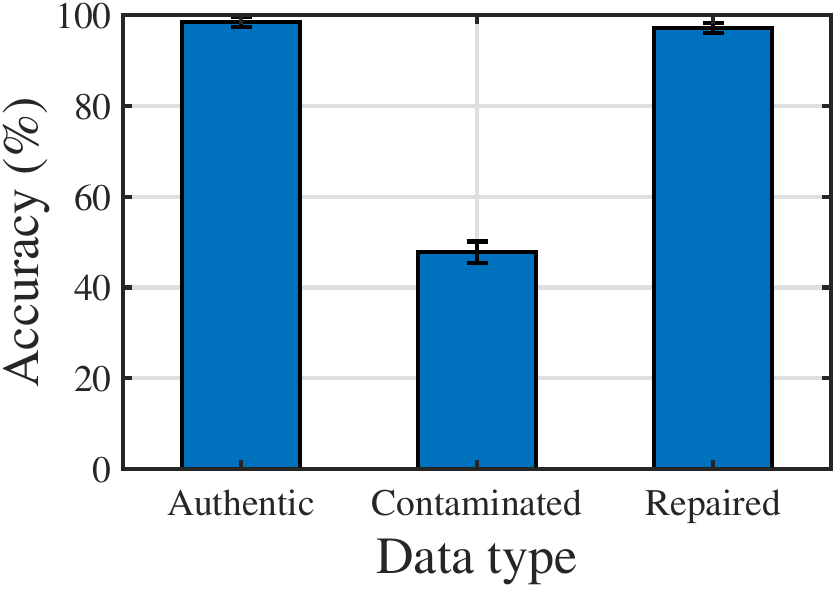}
\label{fig:overallAcc2}
\end{minipage}
}
\subfigure[{\scriptsize Accuracy of GoogleNet}]{
\begin{minipage}[t]{0.29\linewidth}
\centering
\includegraphics[width=1\textwidth]{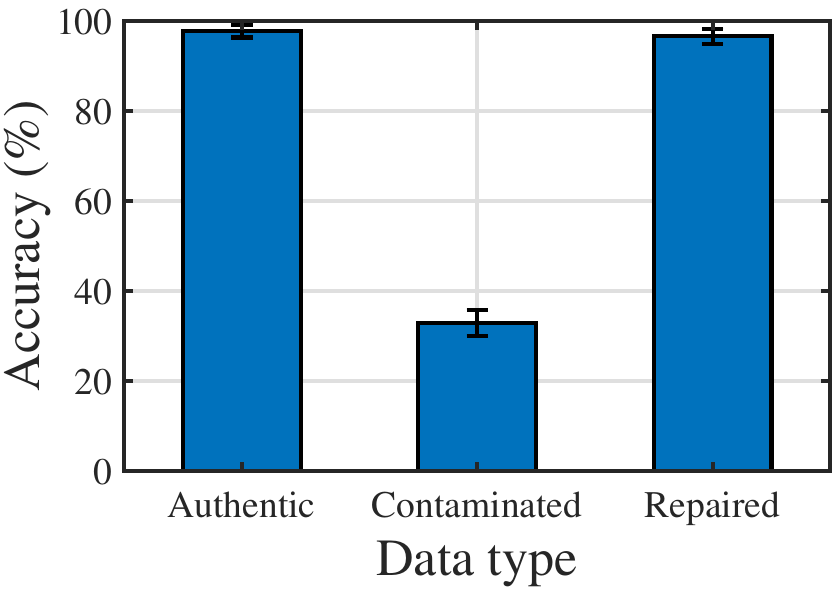}
\label{fig:overallAcc3}
\end{minipage}
}
\subfigure[Precision of YOLO5]{
\begin{minipage}[t]{0.29\linewidth}
\centering
\includegraphics[width=1\textwidth]{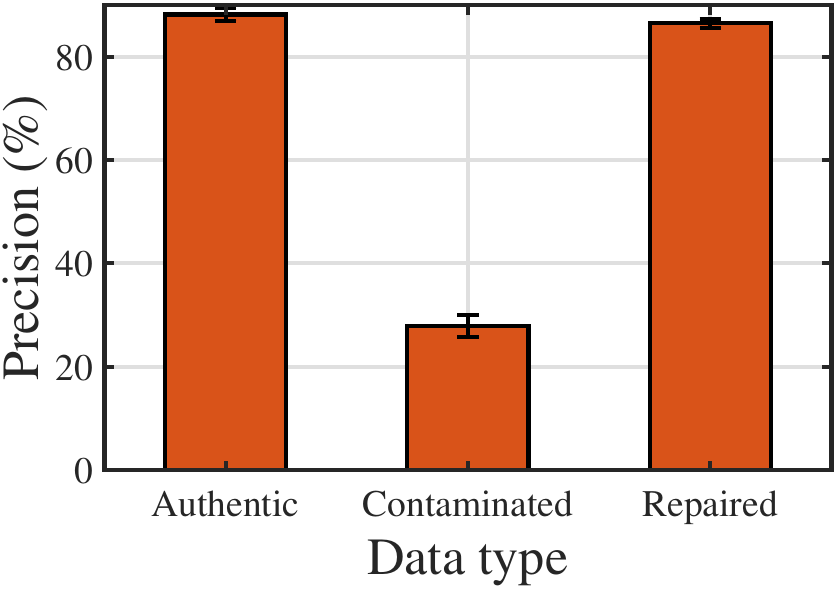}
\label{fig:overallPer1}
\end{minipage}
}
\subfigure[Precision of LeNet]{
\begin{minipage}[t]{0.29\linewidth}
\centering
\includegraphics[width=1\textwidth]{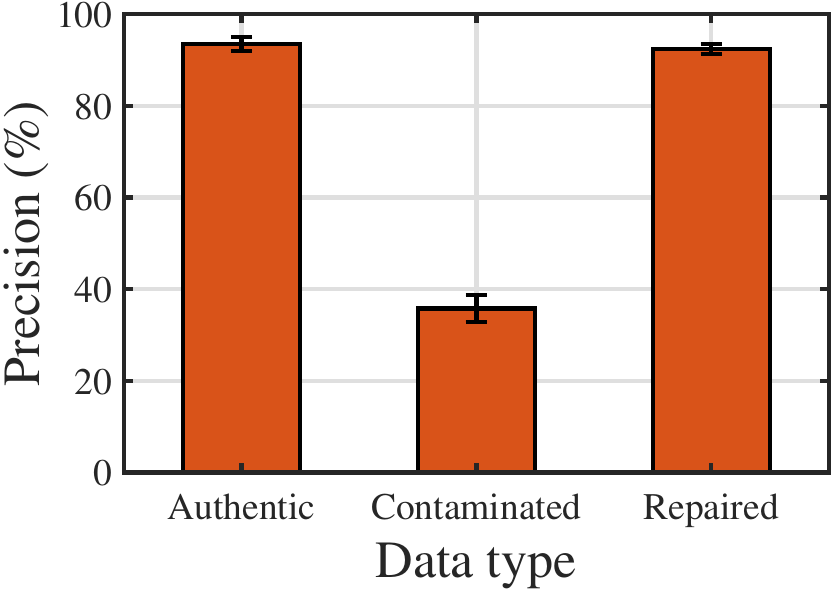}
\label{fig:overallPer2}
\end{minipage}
}
\subfigure[{\scriptsize Precision of GoogleNet}]{
\begin{minipage}[t]{0.29\linewidth}
\centering
\includegraphics[width=1\textwidth]{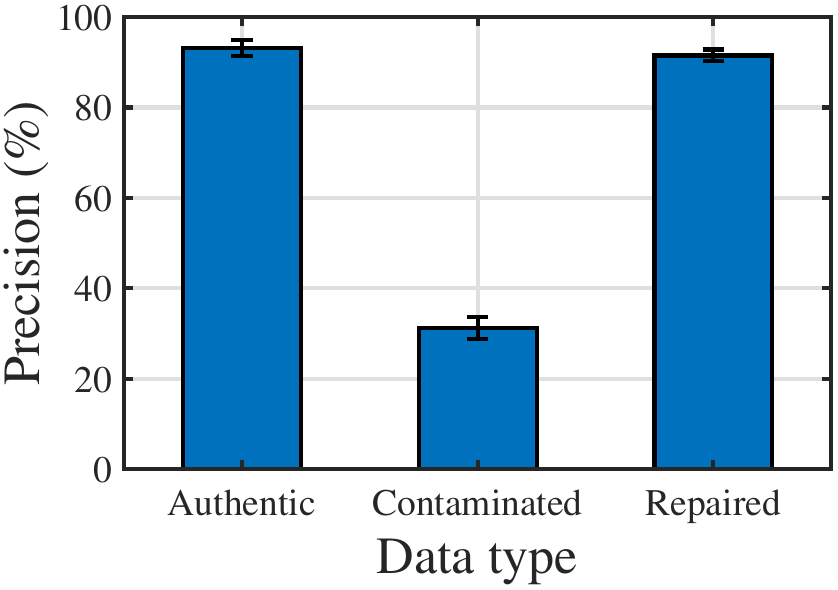}
\label{fig:overallPer3}
\end{minipage}
}
\caption{Feeding three-type sign images into recognition models (i.e., YOLO5, LeNet, and GoogleNet), while presenting the corresponding accuracy and precision distributions in (a)-(c) and (d)-(f) respectively.}
\label{fig:overallPer}
\end{figure}

\subsection{Experiment Result}
\label{subsec:expSetting}
\textbf{Overall performance.}
The goal of \name is to ensure the recognition performance of TSRs by repairing contaminated traffic sign patterns affected by adversarial light patches. In this experiment, to evaluate the effectiveness of the proposed protection mechanism, we feed authentic images, contaminated images, and repaired images into sign recognition models, while comparing the accuracy and precision variations among the three cases. The first three subfigures of Fig.~\ref{fig:overallPer} illustrate the accuracy of the three TSR models when recognizing three types of signs, while the last three show the corresponding precision distribution. The experimental results underscore several critical observations. Firstly, with \name enabled, the average accuracy and precision are 94.7\% and 90.1\%, respectively, representing improvements of 54.8\% and 58.5\% compared to directly inputting contaminated images. Secondly, when recognizing the repaired sign images, TSR provides performance similar to that of the authentic images, with an average difference below 1.4\%. Finally, to recognize the four types (i.e., IS, LL, NLS, PG) of adversarial light patches, the average accuracy and precision variation of three TSRs is less than 1.5\%. These key observations demonstrate that \name can be effectively adapted to common TSRs, skillfully mitigating the interference of light patches and thereby ensuring the accurate recognition of traffic sign images.

\begin{figure}[b]
\centering
\subfigure[One type]{
\begin{minipage}[t]{0.46\linewidth}
\centering
\includegraphics[width=1\textwidth]{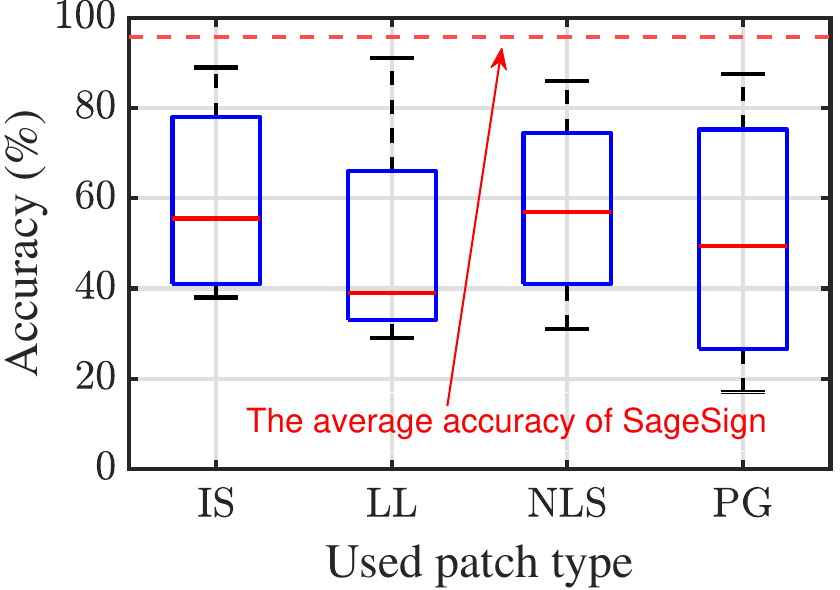}
\label{fig:adT1}
\end{minipage}
}
\subfigure[Four types]{
\begin{minipage}[t]{0.46\linewidth}
\centering
\includegraphics[width=1\textwidth]{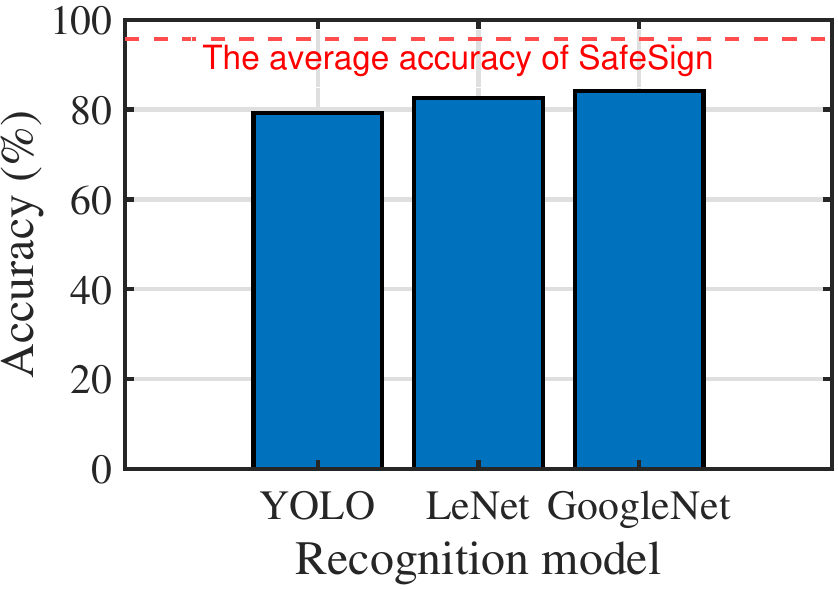}
\label{fig:adT2}
\end{minipage}
}
\caption{Sign recognition performance as (a) adding each type of malicious patches individually, and (b) simultaneously using all types in the model training process.}
\label{fig:adTraining}
\end{figure}

\textbf{The limitation of adversarial training.}
Adversarial training is a widely adopted defense mechanism against various perturbations~\cite{sato2024invisible,duan2021adversarial,zhong2022shadows,lovisotto2021slap}, including light patch attacks. However, this approach necessitates that attackers master attack settings and obtain perturbation samples in advance, rendering it a defense strategy that is model-specific and lacks universality. We hereby assess the differences in the generalization performance between \name and adversarial training. We begin by introducing four distinct types of light patches (i.e., IS, LL, NLS, and PG) into the training process of the recognition model, respectively, then evaluate the model’s performance when recognizing light patches of all types. As illustrated in Fig.~\ref{fig:adT1}, when the training set includes only one type of light patches, the model's average accuracy is noticeably lower than the performance of \sname. Subsequently, we further incorporate all four types of patches into the training process simultaneously, aiming to provide the model with a more comprehensive understanding of these perturbation patterns. Nevertheless, even under these conditions, Fig.~\ref{fig:adT2} shows that the recognition results still fall short of \sname's performance, with the average difference of 13.8\%. We attribute this outcome to the introduction of malicious patches, which amplify intra-class feature variations of signs while diminishing relative inter-class differences, thereby compromising the model’s ability to accurately classify sign types. This result suggests that while adversarial training can be effective in specific contexts, it reveals obvious limitations when facing diversified attack ways. Moreover, this method of adversarial training contradicts the closed-loop design principle of TSR recognition models, which stands as one of its fundamental shortcomings.

\begin{figure}[t]
\centering
\subfigure[YOLO5]{
\begin{minipage}[t]{0.29\linewidth}
\centering
\includegraphics[width=1\textwidth]{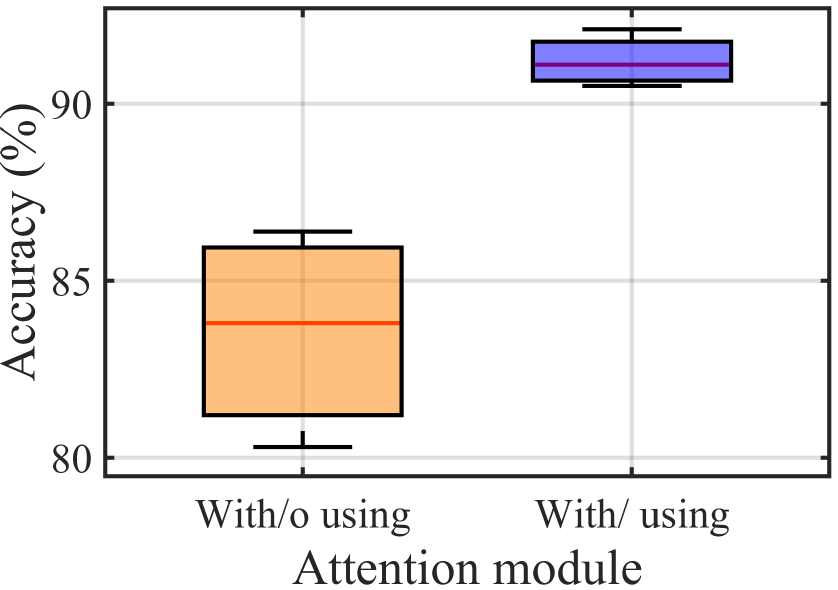}
\label{fig:atten1}
\end{minipage}
}
\subfigure[LeNet]{
\begin{minipage}[t]{0.29\linewidth}
\centering
\includegraphics[width=1\textwidth]{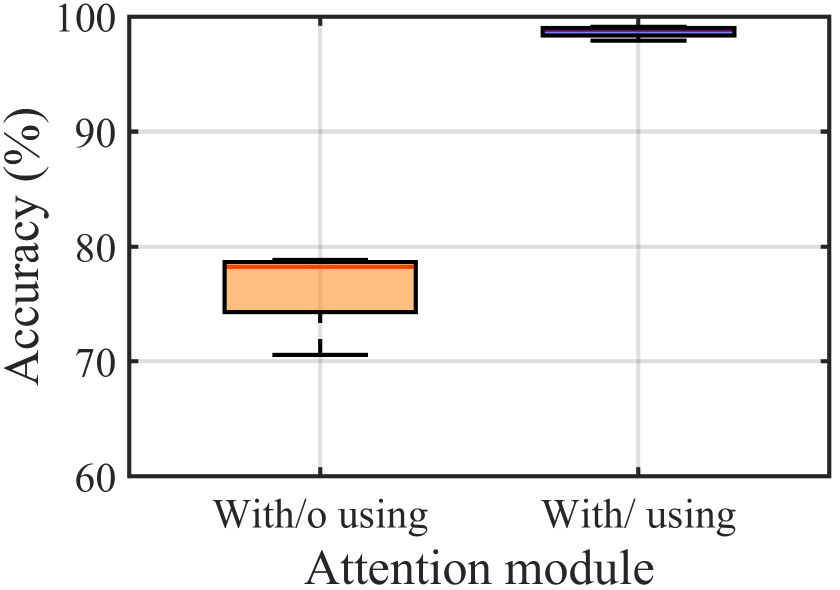}
\label{fig:atten2}
\end{minipage}
}
\subfigure[GoogleNet]{
\begin{minipage}[t]{0.29\linewidth}
\centering
\includegraphics[width=1\textwidth]{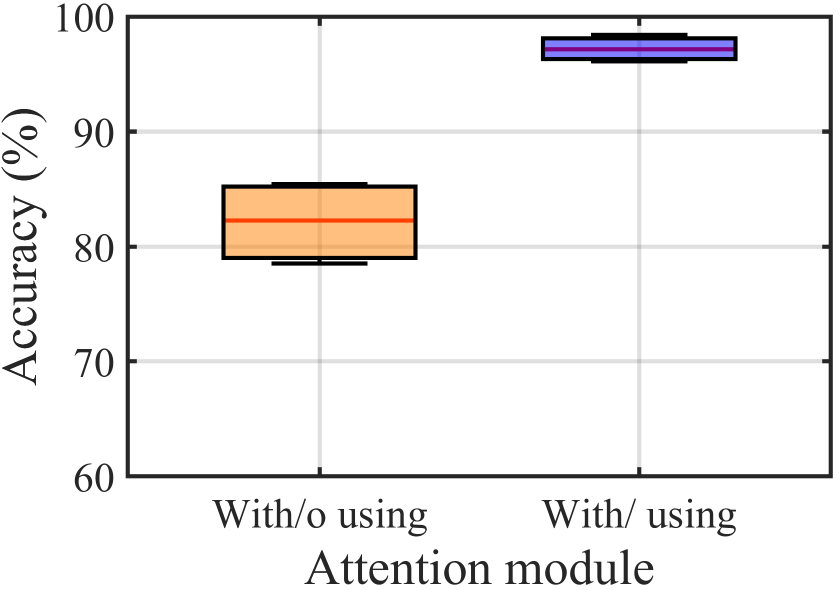}
\label{fig:atten3}
\end{minipage}
}
\caption{Sign recognition performance of (a) YOLO5, (b) LeNet, and (c) GoogleNet, when enabling and disabling the attention module, respectively.}
\label{fig:atten}
\end{figure}

\textbf{The effectiveness of attention mechanism.}
In Section~\ref{sec:imagerec}, we introduce an attention mechanism to harness the abundant information provided by multi-view images, enhancing the ability to repair contaminated signs. To evaluate the effectiveness of this proposed attention module, we perform a comparative analysis, measuring performance with the module both enabled and disabled, while keeping other neural network structures and settings consistent. Fig.~\ref{fig:atten} illustrates the accuracy distribution of recognizing repaired signs by YOLO5, LeNet, and GoogleNet, respectively. The results reveal that enabling the attention mechanism brings an average accuracy improvement of 19.9\%. This significant finding highlights the superior effectiveness of utilizing the attention mechanism, compared to directly inputting multi-view images into the neural network for sign repair.

\begin{figure}[b]
\centering
\begin{minipage}[t]{0.46\linewidth}
\centering
\includegraphics[width=1\textwidth]{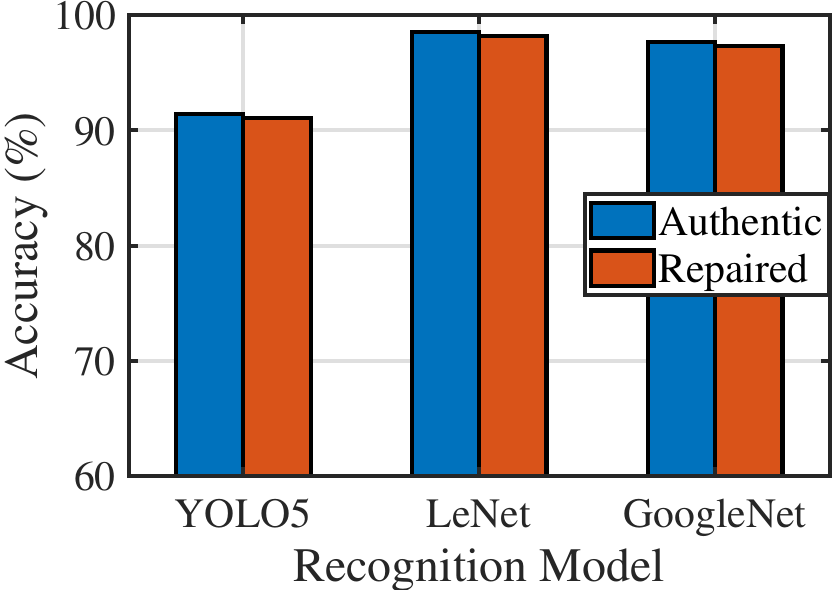}
\caption{Recognition performance as feeding authentic signs and corresponding repaired versions into TSRs, respectively.}
\label{fig:comAcc}
\end{minipage}
\hspace{0.1em}
\begin{minipage}[t]{0.46\linewidth}
\centering
\includegraphics[width=1\textwidth]{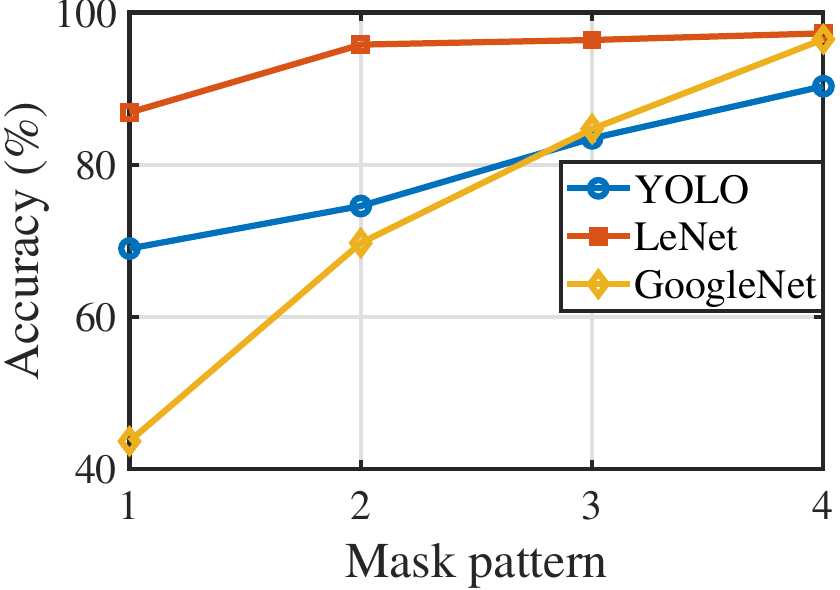}
\caption{Recognition performance as using four different mask patterns.}
\label{fig:accMask}
\end{minipage}
\end{figure}

\textbf{The compatibility with authentic signs.}
In addition to repairing contaminated images, our reconstruction model also needs to ensure compatibility with authentic inputs. To be specific, cameras capture both contaminated and authentic images, which are then directly fed into our repair mechanism. When authentic ones are processed through our reconstruction model, their original sign information that crucial for type recognition should be remained. In this experiment, we input authentic images and their repaired versions into TSRs, observing the recognition performance variation. As shown in Fig.~\ref{fig:comAcc}, the accuracy of TSR models is very similar in both cases, with an average difference of less than 0.5\%. This result indicates that our reconstruction model can effectively preserve the original features of authentic images.

\textbf{The impact of binary mask patterns.}
Binary masks play a crucial role in determining whether the generated contaminated signs effectively cover potential patterns of malicious light patches. With an increase in the variety of mask patterns, the probability that the contaminated samples will precisely cover the area where these malicious patches may emerge, is enhanced. However, given the limitless potential mask patterns, it is impractical to create contaminated images by considering every possible permutation. Fortunately, our preliminary analysis indicates that using four mask patterns (corresponding to using 1, 2, 4, and 6 segmentation lines as described in Section~\ref{sec:noiseGen}) enables \name to effectively repair contaminated images, thereby resisting light patch-based attacks. Fig.~\ref{fig:accMask} presents the recognition performance under four cases: using the first pattern, the first two patterns, the first three patterns, and all four patterns. We can see that there is a consistent improvement in accuracy as the number of patterns increases. When all four masks are simultaneously used, the average accuracy of recognizing the repaired signs approaches that of processing authentic images (i.e., 91.2\%, 98.5\%, and 97.7\% for three recognition models). Therefore, by considering the computation and time cost, we only need to utilize four mask patterns for generating contaminated images in this study.

\begin{figure}[h]
\centering
\begin{minipage}[t]{0.44\linewidth}
\centering
\includegraphics[width=1\textwidth]{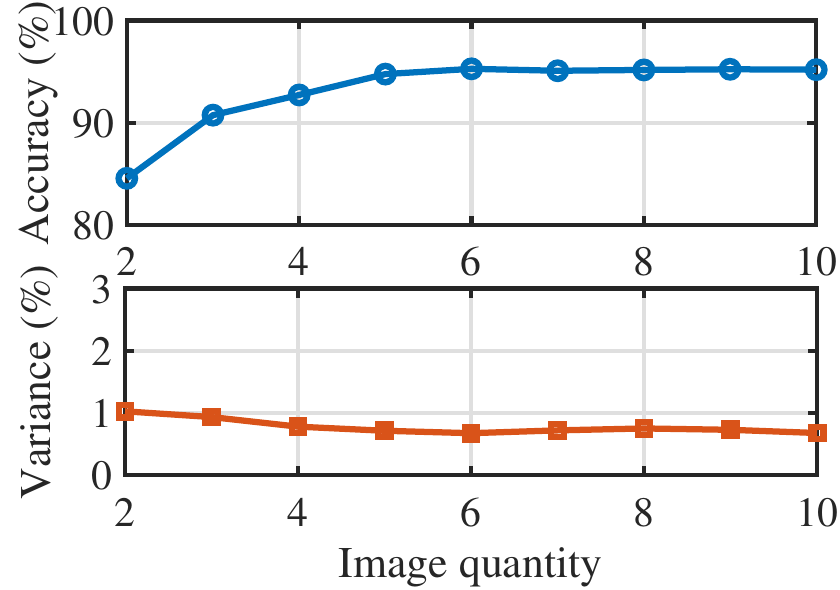}
\caption{Recognition performance as adjusting the multi-view image quantity of each input.}
\label{fig:accQua}
\end{minipage}
\hspace{0.1em}
\begin{minipage}[t]{0.45\linewidth}
\centering
\includegraphics[width=1\textwidth]{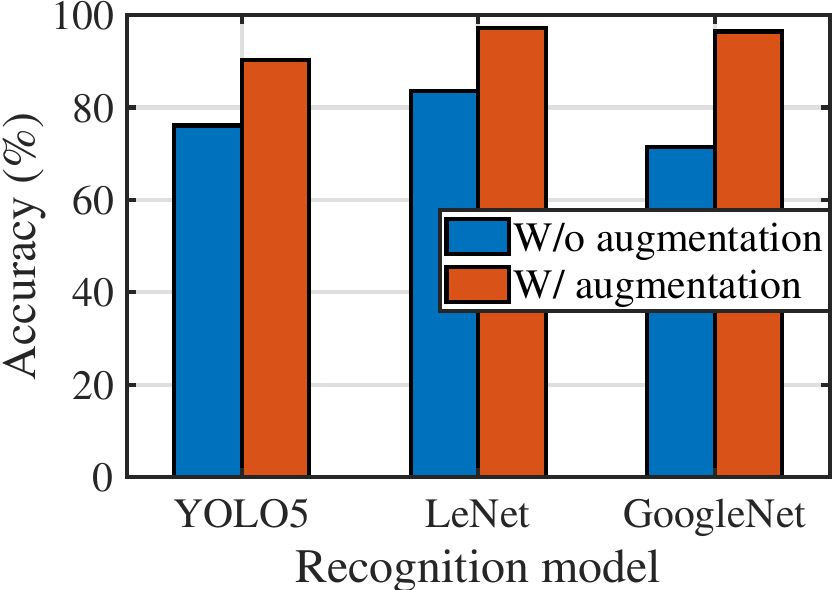}
\caption{Recognition performance as turning on and off the data augmentation module .}
\label{fig:accAug}
\end{minipage}
\end{figure}

\textbf{The impact of inputting multi-view image quantity.}
As described in Section~\ref{sec:imagerec}, images collected from different views contain distinct information and contribute differently to image repair. A crucial parameter to consider is the number of multi-view images used as a single input sample. This parameter setting requires a careful consideration: while more multi-view images enrich the information available, they also demand increased data processing time. For vehicles with limited computational resources, our goal is to achieve satisfactory image repair and traffic sign recognition performance by using the fewest possible images. In this experiment, we adjust the number of multi-view images from two to ten and record the corresponding recognition accuracy. As shown in Fig.~\ref{fig:accQua}, when the number of images exceeds six, the average accuracy of the three TSRs stabilizes around 95\%. Additionally, the accuracy variance for recognizing four types of contaminated signs remains below 0.8\%. This result indicates that an input containing six multi-view images provides sufficient information for sign repair.

\textbf{The impact of data augmentation.}
Providing ample data to a learning-based model is essential for performance optimization. Although the utilized datasets contain tens of thousands of sign images, many types include only a few hundreds of images, which is insufficient for effectively building our contaminated sign generation and reconstruction models. To overcome this issue, we employ image transformation techniques to generate signs under various lighting conditions and shooting settings. In this experiment, we evaluate the performance of recognizing repaired signs both before and after applying data augmentation to the original dataset. As illustrated in Fig.~\ref{fig:accAug}, the average accuracy as enabling the data augmentation module is 94.7\%, with an increase of 17.6\%. The significant improvement in accuracy underscores the substantial positive impact of data augmentation on \sname's construction.

\begin{figure}[t]
\centering
\subfigure[Contaminated sign diversity]{
\begin{minipage}[t]{0.45\linewidth}
\centering
\includegraphics[width=1\textwidth]{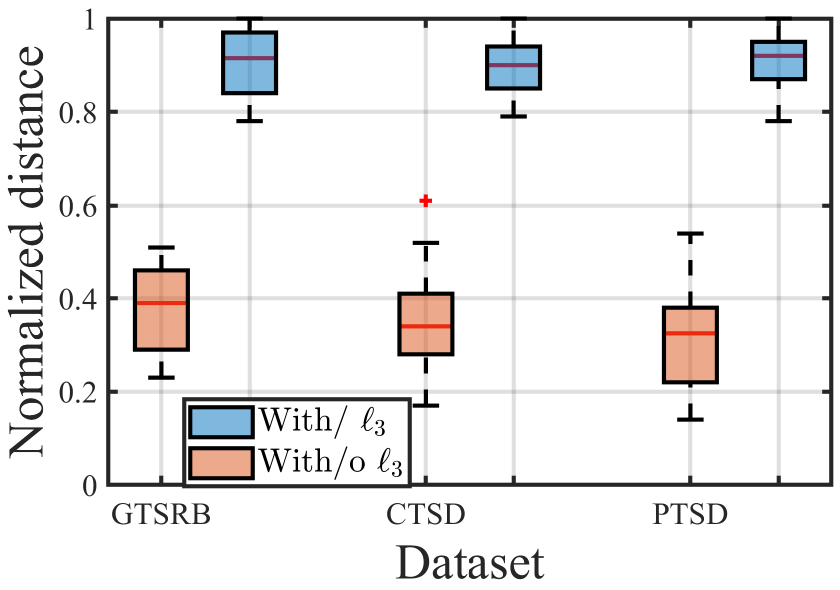}
\label{fig:consignDiff}
\end{minipage}
}
\subfigure[Traffic sign recognition]{
\begin{minipage}[t]{0.46\linewidth}
\centering
\includegraphics[width=1\textwidth]{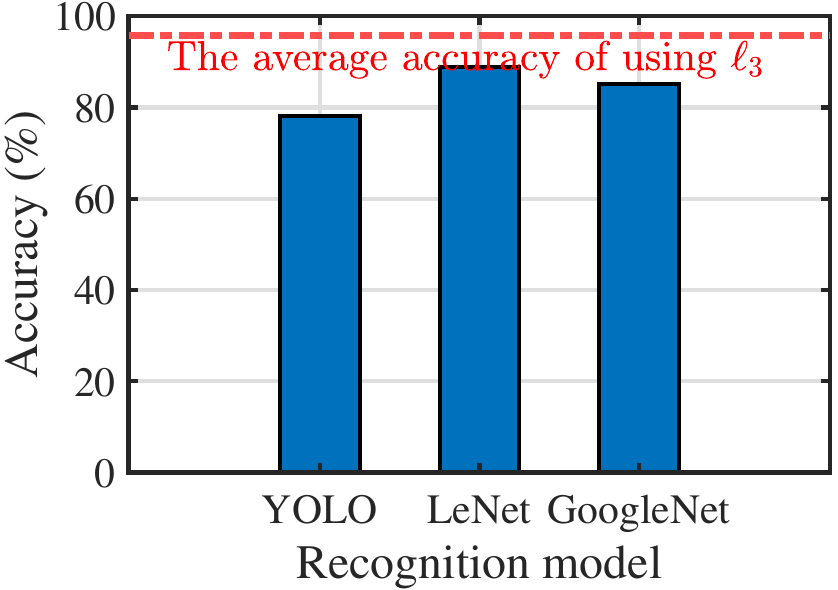}
\label{fig:divPer}
\end{minipage}
}
\caption{Internal difference distributions of multiple contaminated versions of one authentic sign presented in (a) and corresponding recognition performance shown in (b), under without and with using the diversity loss cases.}
\label{fig:maskAcc}
\end{figure}

\textbf{The impact of contaminated sign diversity.} 
In Section~\ref{sec:noiseGen}, we introduce a diversity control loss function (i.e., $\ell_3$) to ensure that the contaminated signs encompass diverse patch patterns. This approach aids the reconstruction model in comprehensively understanding the relationship between authentic and contaminated samples. In this study, we train the model with and without using the diversity loss function, respectively. Subsequently, we calculate the Euclidean distance of contaminated signs for each authentic one and record the classification accuracy in both cases. As illustrated in Fig.~\ref{fig:consignDiff}, the difference among contaminated patterns when using the diversity loss function become large, which indicates better diversity. Furthermore, Fig.\ref{fig:divPer} shows a clear increase in average accuracy of three recognition models, with an improvement of 15.9\%. This result presents the importance of contaminated sign diversity in enhancing the repair capability of the reconstruction model, and demonstrates the effectiveness of our design of diversity loss function.

\section{Discussions and Future Works}
\label{sec:discussion}

\textbf{Extension to multi-sign adversarial attack defense.} Existing light patch-based attacks primarily target single signs (i.e., one sign in each captured image), so we have specifically evaluated our defense mechanism against this emerging threat. However, future attackers may manipulate multiple malicious light sources simultaneously, contaminating several signs within a single image and increasing the misrecognition rate. Current attack methods have not considered this scenario, making it impossible to evaluate \sname's performance under such conditions. Nonetheless, our defense mechanism can be easily extended to effectively counter such multi-sign attacks. This belief is based on the fact that multi-sign attacks essentially expand the disruptive impact of malicious light patches on sign feature patterns. By introducing perturbation noise to multiple sign regions, we can obtain the data of multi-sign adversarial attacks, thereby enabling the reconstruction model to learn the corresponding mapping relationship between contaminated and authentic signs. In the future, we will explore multi-sign attacks and verify the effectiveness of our proposed defense mechanism under these scenarios.

\textbf{Dynamic neural network-driven reconstruction model.}
We utilize the dataset encompassing sign categories of a few regions and countries for building our reconstruction model. In the future, designing a reconstruction model capable of repairing contaminated sign patterns of more types across countries is critical, which always requires deepening the neural network's structure to ensure high accuracy. However, the deeper network often comes with extremely high computation and time costs, thus inapplicable to resource-constrained vehicles in real-time perception environments. Fortunately, researchers in the field of deep learning have increasingly turned their attention to enhancing the model's computation efficiency while maintaining its accuracy. Notably, the recently proposed dynamic neural network~\cite{han2021dynamic} can optimize inference efficiency by dynamically determining which subset of the network architecture to utilize for a given input. By using only a small portion of the entire architecture during inference, this method achieves higher efficiency per unit cost. Consequently, \name can leverage this dynamic architecture to customize the current framework, refining services tailored to different needs of application ranges.

\section{Conclusion}
\label{sec:conclusion}
In this paper, we have focused on building an image inpainting mechanism, \sname, to address the interference caused by lighting patches on traffic sign patterns and recognition models. To provide sufficient data for model training, we leverage image transformation to augment authentic signs. Additionally, we design an adversarial sign generation model to generate various contaminated patterns. Ultimately, with ample authentic and contaminated sign pairs, we build an effective image reconstruction model relying on an attention mechanism-driven multi-view image fusion framework. The superiority of \name lies in its function as a pre-set module that operates at the algorithm level, without retraining the recognition model and modifying hardware architecture. In general, \name enabling sign recognition models effectively counter newly emerging light patch-based attacks, and as a complement to current defense mechanisms, it is crucial for building a secure TSR system.

\bibliographystyle{IEEEtran}
\bibliography{reference}

\end{document}